\documentclass{article}

\usepackage{microtype}
\usepackage{graphicx}
\usepackage{subfigure}
\usepackage{booktabs} 

\usepackage{hyperref}
\usepackage{amssymb, mathtools, amsmath} 
\usepackage[artemisia]{textgreek} 
\usepackage{enumerate, enumitem} 
\usepackage{wasysym}
\DeclareMathOperator*{\argmax}{argmax}  

\usepackage[dvipsnames]{xcolor} 

\usepackage[utf8]{inputenc} 
\usepackage{kotex}

\usepackage[accepted]{icml2020}

\icmltitlerunning{Semi-supervised Disentanglement with Independent Vector Variational Autoencoders}

\begin{document}

\twocolumn[

\icmltitle{Semi-supervised Disentanglement\\with Independent Vector Variational Autoencoders}

\begin{icmlauthorlist}
\icmlauthor{Bo-Kyeong Kim}{kaist}
\icmlauthor{Sungjin Park}{kaist}
\icmlauthor{Geonmin Kim}{kaist}
\icmlauthor{Soo-Young Lee}{kaist,ai}
\end{icmlauthorlist}

\icmlaffiliation{kaist}{School of Electrical Engineering, Korea Advanced Institute of Science and Technology (KAIST), Korea}
\icmlaffiliation{ai}{Institute for Artificial Intelligence, KAIST, Korea}

\icmlcorrespondingauthor{Soo-Young Lee}{sylee@kaist.ac.kr}

\icmlkeywords{Disentanglement Learning, Disentangled Representations, Variational Autoencoder, Semi-supervised Learning, Vector Independence}
\vskip 0.3in
]

\printAffiliationsAndNotice{}


\begin{abstract}

We aim to separate the generative factors of data into two latent vectors in a variational autoencoder. One vector captures class factors relevant to target classification tasks, while the other vector captures style factors relevant to the remaining information. To learn the discrete class features, we introduce supervision using a small amount of labeled data, which can simply yet effectively reduce the effort required for hyperparameter tuning performed in existing unsupervised methods. Furthermore, we introduce a learning objective to encourage statistical independence between the vectors. We show that (i) this vector independence term exists within the result obtained on decomposing the evidence lower bound with multiple latent vectors, and (ii) encouraging such independence along with reducing the total correlation within the vectors enhances disentanglement performance. Experiments conducted on several image datasets demonstrate that the disentanglement achieved via our method can improve classification performance and generation controllability.

\end{abstract}


\section{Introduction} \label{sect1_intro}

A desirably disentangled representation contains individual units, each corresponding to a single generative factor of data while being invariant to changes in other units \cite{tpami13_bengio_reprLearn}. Such interpretable and invariant properties lead to benefits in downstream tasks including image classification and generation \cite{nips14_kingmaDgm, iclrw16_aae, nips17_disent_semiSupDgm, cvpr19_ssl_genEx}.

Variational autoencoders (VAEs) \cite{iclr14_vae} have been actively utilized for unsupervised disentanglement learning \cite{icml18_factorVae, nips18_betaTcVae, aistats19_structDisent, iclr18_dipVae, aistats19_gaoAutoEncTc, nipsw17_bVae_anneal}. To capture the generative factors that are assumed to be statistically independent, many studies have encouraged the independence of latent variables within a representation \cite{iclr17_betaVae, icml18_factorVae, nips18_betaTcVae}. Despite their promising results, the usage of only continuous variables frequently causes difficulty in the discovery of discrete factors (e.g., object categories).

\begin{figure}[t]
\begin{center}
\centerline{\includegraphics[width=6.7cm]{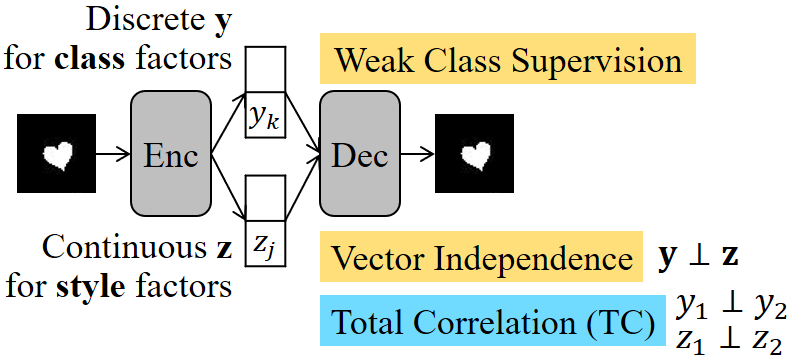}}
\vskip -0.2cm
\caption{Proposed method with discrete and continuous random vectors, $\textbf{y}$ and $\textbf{z}$. Weak classification supervision helps $\textbf{y}$ capture discrete class factors. Our vector independence objective forces $\textbf{y}$ and $\textbf{z}$ to capture different information. Reducing this objective along with the TC improves disentanglement.}
\label{figVecIdp_overview}
\end{center}
\vskip -0.34in
\end{figure}

To address this issue, researchers have utilized discrete variables together with continuous variables to separately capture discrete and continuous factors \cite{nips18_jointVae, nips14_kingmaDgm} and trained their models by maximizing the evidence lower bound (ELBO). In this paper, we show that the ability of disentanglement in their models is derived from not only using the two types of variables but also encouraging several sources of disentanglement. We expose these sources by decomposing the ELBO.

Unsupervised learning with discrete and continuous variables is difficult, because
continuous units with a large informational capacity often store all information, causing discrete units to store nothing and be neglected by models \cite{nips18_jointVae}. Previously, this issue was mostly solved using hyperparameter-sensitive capacity controls \cite{nips18_jointVae} or additional steps for inferring discrete features during training \cite{icml19_cascadeVae}. In contrast, we simply inject weak supervision with a few class labels and effectively resolve the difficulty in learning of discrete variables. \citet{icml18_commAssump} also suggested the exploitation of available supervision for improving disentanglement.

We introduce a semi-supervised disentanglement learning method, which is shown in Figure \ref{figVecIdp_overview}. A VAE was used to extract two feature vectors: one containing discrete variables to capture class factors and the other vector containing continuous variables to capture style factors. The contributions of this work to the relevant field of study are as follows:

\vskip -1.4cm

\begin{itemize}[noitemsep]

\setlength{\leftskip}{-0.3cm}

\item We introduce the vector independence objective that measures the statistical independence between latent vectors in VAEs and enables different vectors to store different information. We name our model an Independent Vector VAE (IV-VAE).

\item We decompose the ELBO containing multiple latent vectors to reveal the vector independence term along with well-known total correlation (TC) terms that measure independence between the variables within the vectors. We show that these terms are the sources of disentanglement in jointly learning discrete and continuous units. 

\item We introduce supervision with a small number of class labels, significantly reducing the difficulty in learning discrete class features.
 
\item We empirically show that our method enhances disentanglement learning on several image datasets, positively affecting the classification and controllable generation.

\setlength{\leftskip}{0pt}

\end{itemize}

\vskip -0.2cm

The supplementary material includes derivation details and additional results, and the sections are indicated with S (e.g., Section S1).


\section{Related Work: Disentanglement Learning with VAEs} \label{sect2_background}

A $J$-dimensional random vector, $\textbf{z}=[z_1,...,z_J]^T$, consists of scalar-valued random variables, $\{z_j\}$. Previous methods have encouraged independence between variables, whereas our method encourages independence between vectors (by introducing another vector, $\textbf{y}$) along with independence between the variables within the vectors.

\subsection{Promoting Independence of Continuous Variables}
\label{sect_2bg_1_IndepConti}

Given dataset $D = \{ \textbf{x}^{(1)},...,\textbf{x}^{(N)} \}$ containing $N$ i.i.d. samples of random vector \textbf{x}, a VAE learns latent vector $\textbf{z}$ involved in the data generation process. The VAE consists of encoder $q_{\phi}(\textbf{z}|\textbf{x}^{(n)})$ and decoder $p_{\theta}(\textbf{x}^{(n)}|\textbf{z})$ trained by maximizing the ELBO on $\mathbb{E}_{q(\textbf{x})} \big[\log p_{\theta}(\textbf{x}^{(n)}) \big]$, where the empirical distribution is represented as $q(\textbf{x}) = \frac{1}{N} \sum\nolimits_{n=1}^{N}\delta(\textbf{x}-\textbf{x}^{(\textit{n})})$. The training objective consists of the reconstruction term and the KL divergence from the prior to individual posteriors:

\vskip -0.4cm

\begin{equation} \label{eqn_betavae}
\mathcal{L}^z_{\beta} = \mathbb{E}_{q(\textbf{x})} \Big[ \mathbb{E}_{q_{\phi}(\textbf{z}|\textbf{x})} \big[ \log  p_{\theta}(\textbf{x}|\textbf{z}) \big]  \Big] - \beta\mathcal{L}^z_{KL} 
\end{equation}

\vskip -0.4cm

\begin{equation} \label{eqn_vae_kl}
\mathcal{L}^z_{KL} = \mathbb{E}_{q(\textbf{x})} \Big[ \textrm{D}_{\textrm{KL}}\big( q_{\phi}(\textbf{z}|\textbf{x}) || p(\textbf{z})\big) \Big],
\end{equation}

\vskip -0.1cm

\noindent where $\beta=1$ and $\beta>1$ in (\ref{eqn_betavae}) represent the vanilla VAE and $\beta$-VAE \cite{iclr17_betaVae} objectives, respectively. Under factorized prior $p(\textbf{z})=\prod\nolimits_{j} p(z_{j})$ (e.g., $N(\textbf{0},\textbf{I})$), the $\beta$-VAE enhances the independence of latent variables, leading to disentanglement.

To understand the disentangling mechanism in VAEs, the KL term (\ref{eqn_vae_kl}) was decomposed into (\ref{eqn_decomp_hoffman}) in \cite{nipsw16_elboSurgery} and further into (\ref{eqn_decomp_chen}) under the factorized prior in \cite{nips18_betaTcVae}:

\vskip -0.42cm
\begin{equation} \label{eqn_decomp_hoffman}
\mathcal{L}^z_{KL} = I_{q_{\phi}}(\textbf{z};\textbf{x}) +\textrm{D}_{\textrm{KL}}\big( q_{\phi}(\textbf{z}) || p(\textbf{z})\big)
\end{equation}

\vskip -0.63cm

\begin{equation} \label{eqn_decomp_chen}
\mathcal{L}^z_{KL} = I_{q_{\phi}}(\textbf{z};\textbf{x}) +\textrm{D}_{\textrm{KL}} \Big( q_{\phi}(\textbf{z}) || \prod\nolimits_{j} q_{\phi}(z_{j}) \Big) +\mathcal{L}^z_{Reg}, 
\end{equation}

\vskip -0.22cm

\noindent where $q_{\phi}(\textbf{z}) = \int_{\textbf{x}}q_{\phi}(\textbf{z}|\textbf{x})q(\textbf{x})d\textbf{x} = \frac{1}{N} \sum\nolimits_{n=1}^{N}q_{\phi}(\textbf{z}|\textbf{x}^{(\textit{n})})$ is the \textit{aggregate} posterior \cite{iclrw16_aae} that describes the latent structure for all data samples, $I_{q_{\phi}}(\textbf{z};\textbf{x})=\textrm{D}_{\textrm{KL}}\big( q_{\phi}(\textbf{z},\textbf{x}) || q_{\phi}(\textbf{z})q(\textbf{x})\big)$ is the mutual information (MI) between the data and latent vectors under empirical distribution $q_{\phi}(\textbf{z}|\textbf{x})q(\textbf{x})$, and $\mathcal{L}^z_{Reg} = \sum\nolimits_{j}\textrm{D}_{\textrm{KL}} \big( q_{\phi}(z_j)||p(z_j)\big)$ is the regularization term for penalizing individual latent dimensions that excessively differ from their priors.

The second term in (\ref{eqn_decomp_chen}) represents the TC \cite{ibm1960_tc} that measures the statistical dependency of more than two random variables. \citet{icml18_factorVae} and \citet{nips18_betaTcVae} argued that (i) $\beta$-VAEs heavily penalize the MI term, $I_{q_{\phi}}(\textbf{z};\textbf{x})$, along with the other terms, causing \textbf{z} to be less informative about \textbf{x}, and (ii) the TC term is the source of disentanglement and thus should be strongly penalized. To address these issues, \citet{icml18_factorVae} proposed the FactorVAE by adding the TC term to the vanilla VAE objective, and \citet{nips18_betaTcVae} proposed the $\beta$-TCVAE by separately controlling the three terms in (\ref{eqn_decomp_chen}) using individual weights.

The previous methods \cite{iclr17_betaVae, icml18_factorVae, nips18_betaTcVae} learned disentangled representations by encouraging the independence of continuous variables. They primarily employed Gaussian distributions for variational posteriors and consequently focused on modeling continuous generative factors. However, most datasets naturally contain discrete factors of variation (e.g., object classes), which these methods frequently fail to capture. We address this issue by incorporating discrete random variables and further enhance the disentanglement by promoting the independence between one set of discrete variables and another set of continuous variables.

\subsection{Utilizing Discrete and Continuous Variables}
\label{sect_2bg_2_DiscConti}

To separately capture discrete and continuous factors, researchers have proposed the simultaneous utilization of discrete and continuous variables, which are stored in two latent vectors, \textbf{y} and \textbf{z}, respectively. By introducing joint prior $p(\textbf{y},\textbf{z})$, approximate posterior $q_{\phi}(\textbf{y},\textbf{z}|\textbf{x})$, and likelihood $p_{\theta}(\textbf{x}|\textbf{y},\textbf{z})$, the ELBO containing the reconstruction and KL terms becomes (see Section S1 for the derivation)

\vskip -1cm

\begin{equation} \label{eqn_elbo_unsup_x}
\mathcal{L}^{y,z}_{\beta} = \mathbb{E}_{q(\textbf{x})} \Big[ \mathbb{E}_{q_{\phi}(\textbf{y},\textbf{z}|\textbf{x})} \big[ \log  p_{\theta}(\textbf{x}|\textbf{y},\textbf{z}) \big]\Big] - \beta
\mathcal{L}^{y,z}_{KL},
\end{equation}

\vskip -0.45cm

\begin{equation} \label{eqn_vae_kl_yz}
\mathcal{L}^{y,z}_{KL} = \mathbb{E}_{q(\textbf{x})} \Big[ \textrm{D}_{\textrm{KL}} \big( q_{\phi}(\textbf{y},\textbf{z} | \textbf{x}) || p(\textbf{y},\textbf{z}) \big) \Big].
\end{equation}

\vskip -0.15cm

By assuming factorized prior $p(\textbf{y},\textbf{z})=p(\textbf{y})p(\textbf{z})$, the KL can be decomposed depending on the factorized form of $q_{\phi}(\textbf{y},\textbf{z}|\textbf{x})$; \citet{nips14_kingmaDgm} assumed $q_{\phi}(\textbf{y}|\textbf{x})q_{\phi}(\textbf{z}|\textbf{y},\textbf{x})$\footnote{The unsupervised objective in Eqn. 7 of \citet{nips14_kingmaDgm} can be reformulated as $\mathcal{L}^{y,z}_{KL}  = \mathbb{E}_{q(\textbf{x})} \Big[ \textrm{D}_{\textrm{KL}}\big( q_{\phi}(\textbf{y}|\textbf{x}) || p(\textbf{y})\big)  +  \mathbb{E}_{q_{\phi}(\textbf{y}|\textbf{x})} \big[ \textrm{D}_{\textrm{KL}}\big( q_{\phi}(\textbf{z}|\textbf{y},\textbf{x}) || p(\textbf{z})\big)  \big] \Big] $. Eqn. 7 was extended to reveal the objective for labeled data and the entropy for discrete $\textbf{y}$.}, and \citet{nips18_jointVae} assumed $q_{\phi}(\textbf{y}|\textbf{x})q_{\phi}(\textbf{z}|\textbf{x})$ to derive

\vskip -0.63cm

\begin{equation} \label{eqn_elbo_unsup_x_qz_x}
\begin{split}
\mathcal{L}^{y,z}_{KL}  =&\; \mathbb{E}_{q(\textbf{x})} \Big[ \textrm{D}_{\textrm{KL}}\big( q_{\phi}(\textbf{y}|\textbf{x}) || p(\textbf{y})\big) \\ & + \textrm{D}_{\textrm{KL}}\big( q_{\phi}(\textbf{z}|\textbf{x}) || p(\textbf{z})\big)  \Big] 
\end{split}.
\end{equation}

\vskip -0.3cm

In contrast, we decompose the KL term (\ref{eqn_vae_kl_yz}) to reveal the existence of the vector independence term between $\textbf{y}$ and $\textbf{z}$ as well as the TC terms within the vectors (see our decomposition in (\ref{eqn_decompElbo})). Moreover, our method explicitly encourages the vector independence term while not penalizing the latent-data MI terms, $I_{q_{\phi}}(\textbf{y};\textbf{x})$ and $I_{q_{\phi}}(\textbf{z};\textbf{x})$, to obtain disentangled informative features. We show that our method outperforms the $\beta$-VAE \cite{iclr17_betaVae} and its variant \cite{nips18_jointVae}, which strongly penalize the KL term (\ref{eqn_elbo_unsup_x_qz_x}) and consequently minimize the latent-data MI terms.

In addition, unsupervised learning with discrete and continuous variables often causes the discrete units to capture less information and be discarded by the model \cite{nips18_jointVae} because of a larger informational capacity of continuous units than that of discrete units. To address this issue, existing unsupervised methods involve sophisticated hyperparameter tunings or additional computations. For example, \citet{nips18_jointVae} modified (\ref{eqn_elbo_unsup_x_qz_x}) using capacity control terms separately for discrete and continuous units, and \citet{icml19_cascadeVae} proposed an alternating optimization between inferring probable discrete features and updating the encoder. In contrast, we introduce weak classification supervision to guide the encoder to store class factors in discrete units; this method simply but significantly reduces the effort needed for designing inference steps.


\section{Proposed Approach} \label{sect3_method}

The problem scenario is identical to that described in Section \ref{sect_2bg_2_DiscConti}, with discrete random vector $\textbf{y}\in{\rm I\!R}^{K}$ for capturing class factors and continuous random vector $\textbf{z}\in{\rm I\!R}^{J}$ for capturing style factors. First, we show that the ELBO can be decomposed into the proposed vector independence term and the others. Then, we present our semi-supervised learning (SSL) strategy.

\subsection{Learning of Independent Latent Vectors} \label{sect_3mtd_1_vecIdp}

\subsubsection{Vector Independence Objective}

Assuming the conditional independence of $q_{\phi}(\textbf{y},\textbf{z}|\textbf{x})=q_{\phi}(\textbf{y}|\textbf{x})q_{\phi}(\textbf{z}|\textbf{x})$, an encoder produces the parameters for variational posteriors $q_{\phi}(\textbf{y}|\textbf{x}^{(\textit{n})})$ and $q_{\phi}(\textbf{z}|\textbf{x}^{(\textit{n})})$, for the \textit{n}-th sample, $\textbf{x}^{(\textit{n})}$. The aggregate posteriors that capture the entire latent space under data distribution $q(\textbf{x})$ are defined as $q_{\phi}(\textbf{z}) = \frac{1}{N} \sum\nolimits_{n=1}^{N}q_{\phi}(\textbf{z}|\textbf{x}^{(\textit{n})})$, $q_{\phi}(\textbf{y}) = \frac{1}{N} \sum\nolimits_{n=1}^{N}q_{\phi}(\textbf{y}|\textbf{x}^{(\textit{n})})$, and

\vskip -0.7cm

\begin{equation} \label{eqn_qzy}
q_{\phi}(\textbf{y},\textbf{z})=\mathbb{E}_{q(\textbf{x})} \big[ q_{\phi}(\textbf{y},\textbf{z}|\textbf{x}) \big] = \frac{1}{N} \sum_{n=1}^{N}q_{\phi}(\textbf{y},\textbf{z}|\textbf{x}^{(\textit{n})}),
\end{equation}

\vskip -0.3cm

\noindent where $q_{\phi}(\textbf{y},\textbf{z}|\textbf{x})$ is computed using its decomposition form.\footnote{In this paper, we assumed the conditional independence by following \cite{nips18_jointVae}. However, $q_{\phi}(\textbf{y},\textbf{z}|\textbf{x})$ can be computed as $q_{\phi}(\textbf{y}|\textbf{x})q_{\phi}(\textbf{z}|\textbf{y},\textbf{x})$ by incorporating architectural dependency from $\textbf{y}$ to $\textbf{z}$. In this case, only $q_{\phi}(\textbf{z}) = \frac{1}{N} \sum\nolimits_{n=1}^{N} \mathbb{E}_{q_{\phi}(\textbf{y}|\textbf{x}^{(\textit{n})})} \big[ q_{\phi}(\textbf{z}|\textbf{y},\textbf{x}^{(\textit{n})})\big] $ is revised, whereas the computation of (\ref{eqn_qzy}) and (\ref{eqn_vecIdp}) remains unchanged.} Then, we define our \textit{vector independence} objective as the MI between the two vectors that measures their statistical dependency:

\vskip -0.3cm

\begin{equation} \label{eqn_vecIdp}
\mathcal{L}^{y,z}_{VecIdp} = \textrm{D}_{\textrm{KL}}\big( q_{\phi}(\textbf{y},\textbf{z}) || q_{\phi}(\textbf{y})q_{\phi}(\textbf{z})\big).
\end{equation}

\vskip -0.05cm

\noindent Thr reduction of this term can enforce \textbf{y} and \textbf{z} to capture different semantics. Here, we emphasize that our method is applicable to cases with multiple \textit{L} latent vectors by extending (\ref{eqn_vecIdp}) to $\textrm{D}_{\textrm{KL}}\big( q_{\phi}(\textbf{z}_{1},...,\textbf{z}_{L}) || \prod\nolimits_{l=1}^{L} q_{\phi}(\textbf{z}_{l})\big)$, which has a form similar to the TC computed over the variables (i.e., $\textrm{D}_{\textrm{KL}} \big( q_{\phi}(\textbf{z}) || \prod\nolimits_{j=1}^{J} q_{\phi}(z_{j})\big)$) but is computed over the vectors. The relationship between the TC and the vector independence term is similar to that between the objectives of independent component analysis (ICA; \citet{sp91_ica_1, nips96_ica_2_amari}) and independent vector analysis (IVA; \citet{iva_1_conf, iva_2_jour}).

\subsubsection{Our ELBO Decomposition}

\citet{nips18_betaTcVae} showed that the TC term measuring the dependency between latent variables (i) exists in the decomposion of the ELBO containing a single latent vector, $\textbf{z}$, and (ii) is a source of disentanglement in VAEs. Similarly, we reveal that the vector independence term (i) exists in the decomposition of the ELBO containing two latent vectors, \textbf{y} and \textbf{z}, and (ii) is \textit{another} source of disentanglement.

Concretely, we decompose the KL term (\ref{eqn_vae_kl_yz}) of the ELBO
into (\ref{eqn_decompElbo_pre}) under $p(\textbf{y},\textbf{z})=p(\textbf{y})p(\textbf{z})$ and further into (\ref{eqn_decompElbo}) under $p(\textbf{y})=\prod\nolimits_{k} p(y_{k})$ and $p(\textbf{z})=\prod\nolimits_{j} p(z_{j})$:

\vskip -0.3cm

\begin{equation} \label{eqn_decompElbo_pre}
\begin{split}
 \mathcal{L}^{y,z}_{KL} =&\;  I_{q_{\phi}}(\textbf{y},\textbf{z};\textbf{x})
+  \textrm{D}_{\textrm{KL}}\big( q_{\phi}(\textbf{y},\textbf{z})||q_{\phi}(\textbf{y})q_{\phi}(\textbf{z})\big)\\& + \textrm{D}_{\textrm{KL}} \Big( q_{\phi}(\textbf{y}) || p(\textbf{y}) \Big)  +\textrm{D}_{\textrm{KL}} \Big( q_{\phi}(\textbf{z}) || p(\textbf{z}) \Big) 
 \end{split}
\end{equation}

\vskip -0.55cm

\begin{equation} \label{eqn_decompElbo}
\begin{split}
 \mathcal{L}^{y,z}_{KL} =&\;  I_{q_{\phi}}(\textbf{y},\textbf{z};\textbf{x})
+  \textrm{D}_{\textrm{KL}}\big( q_{\phi}(\textbf{y},\textbf{z})||q_{\phi}(\textbf{y})q_{\phi}(\textbf{z})\big)\\& +  \textrm{D}_{\textrm{KL}} \Big( q_{\phi}(\textbf{y}) || \prod\nolimits_{k} q_{\phi}(y_{k}) \Big)  +  \mathcal{L}^y_{Reg} \\& +\textrm{D}_{\textrm{KL}} \Big( q_{\phi}(\textbf{z}) || \prod\nolimits_{j} q_{\phi}(z_{j}) \Big)  +\mathcal{L}^z_{Reg}  
\end{split},
\end{equation}

\vskip -0.3cm

\noindent where $ I_{q_{\phi}}(\textbf{y},\textbf{z};\textbf{x}) = \textrm{D}_{\textrm{KL}}\big( q_{\phi}(\textbf{y},\textbf{z},\textbf{x})||q_{\phi}(\textbf{y},\textbf{z})q(\textbf{x}) \big) $ is the MI between \textbf{x} and latent vectors $\textbf{y}$ and $\textbf{z}$, and $\mathcal{L}^y_{Reg} = \sum\nolimits_{k}\textrm{D}_{\textrm{KL}} \big( q_{\phi}(y_k)||p(y_k)\big)$ and $\mathcal{L}^z_{Reg} = \sum\nolimits_{j}\textrm{D}_{\textrm{KL}} \big( q_{\phi}(z_j)||p(z_j)\big)$ are the dimension-wise regularization terms. The derivation from (\ref{eqn_decompElbo_pre}) to (\ref{eqn_decompElbo}) is motivated by that from (\ref{eqn_decomp_hoffman}) to (\ref{eqn_decomp_chen}). See Section S2 for the derivation details.

As suggested by \citet{icml18_factorVae} and \citet{nips18_betaTcVae}, data-latent MI $I_{q_{\phi}}(\textbf{y},\textbf{z};\textbf{x})$ is not penalized during training so as to allow the latent vectors to capture data information. The \textit{second} term in the RHS is our vector independence objective, and the \textit{third} and \textit{fifth} terms are the TC terms for the  variables in $\textbf{y}$ and $\textbf{z}$, respectively. We empirically show that simultaneously reducing these three terms provides better disentanglement compared to penalizing only the TC terms without considering the vector independence. The regularization terms, $\mathcal{L}^y_{Reg}$ and $\mathcal{L}^z_{Reg}$, forbid individual latent variables from deviating largely from the priors. 

A concurrent work \cite{aistats19_structDisent} introduced a decomposition similar to our result in (\ref{eqn_decompElbo}). Their derivation was initiated by augmenting the ELBO with a data entropy term (i.e., $-\mathbb{E}_{q(\textbf{x})} \big[ \log  q(\textbf{x}) \big]$). In contrast to our method, their method with discrete and continuous units uses purely unsupervised learning, which often causes the discrete units to be ignored by the model.

\subsubsection{Relationship between the Vector Independence Objective and TC}

Here, we investigate the relationship between vector independence objective $\mathcal{L}^{y,z}_{VecIdp}$ (\ref{eqn_vecIdp}) and the following TC terms:

\vskip -0.55cm

\begin{equation} \label{eqn_lossVar_sepa}
\begin{split}
& \mathcal{L}^{y}_{TC}= \textrm{D}_{\textrm{KL}} \Big( q_{\phi}(\textbf{y}) || \prod\nolimits_{k}q_{\phi}(y_{k}) \Big) \quad \textrm{and} \quad \\& \mathcal{L}^{z}_{TC} = \textrm{D}_{\textrm{KL}} \Big( q_{\phi}(\textbf{z}) || \prod\nolimits_{j} q_{\phi}(z_{j})\Big) 
\end{split}
\end{equation}

\vskip -0.7cm

\begin{equation} \label{eqn_lossVar_collec}
\mathcal{L}^{y,z}_{TC}= \textrm{D}_{\textrm{KL}} \Big( q_{\phi}(\textbf{y}, \textbf{z}) || \prod\nolimits_{k}q_{\phi}(y_{k}) \prod\nolimits_{j}q_{\phi}(z_{j})\Big).
\end{equation}

\vskip -0.3cm

\noindent (\ref{eqn_lossVar_sepa}) measures the independence of the variables \textit{within} each vector (hereafter called the “separate” TC). In addition, (\ref{eqn_lossVar_collec}) \textit{simultaneously} considers the variables in $\textbf{y}$ and $\textbf{z}$ (hereafter called the “collective” TC), and it can be viewed as the TC on concatenated vector $\textbf{h}=[$\textbf{y}$ ; \, $\textbf{z}$]$ (i.e., $ \textrm{D}_{\textrm{KL}} \big( q_{\phi}(\textbf{h}) || \prod\nolimits_{i} q_{\phi}(h_{i}) \big)$).

We introduce two relationships. First, perfectly penalized collective TC indicates perfect vector independence:

\vskip -0.42cm

\begin{equation} \label{relation_1}
\mathcal{L}^{y,z}_{TC}=0 \; \Rightarrow  \; \mathcal{L}^{y}_{TC}=\mathcal{L}^{z}_{TC}=\mathcal{L}^{y,z}_{VecIdp}=0
\end{equation}

\vskip -0.15cm

\noindent by letting $q_{\phi}(\textbf{z}) = \prod\nolimits_{j} q_{\phi}(z_{j})$ and $q_{\phi}(\textbf{y})=\prod\nolimits_{k} q_{\phi}(y_{k})$. In this case, the vector independence objective would be naturally satisfied, resulting in unnecessary optimization. However, this case is rare because of the existence of other loss terms (e.g., a reconstruction term) that often prevent the collective TC from being zero. Furthermore, under the factorized prior, the perfectly penalized TC may be undesirable because it could imply the occurrence of posterior collapse (i.e., learning a trivial posterior that collapses to the prior and fails to capture data features). We present the experimental setup and results regarding this relationship in Sections S5 and S6.

Second, perfect vector independence does not ensure that all variables within and between the vectors are perfectly independent, i.e., $ \mathcal{L}^{y,z}_{VecIdp}=0 \;  \nRightarrow \; \mathcal{L}^{y}_{TC}=\mathcal{L}^{z}_{TC}=\mathcal{L}^{y,z}_{TC}=0$. However, perfect vector independence ensures that the collective TC is the sum of the two separate TCs, i.e., $\mathcal{L}^{y,z}_{VecIdp}=0 \; \Rightarrow \; \mathcal{L}^{y,z}_{TC}=\mathcal{L}^{y}_{TC}+\mathcal{L}^{z}_{TC}$ (see Section S3 for the derivation).

\subsection{Semi-supervised Learning (SSL)}

A weak classification supervision guides $\textbf{y}$ to suitably represent discrete class factors. In addition, our vector independence objective further enforces $\textbf{y}$ and $\textbf{z}$ to capture different types of information. In our experiments, we simplify the problem setup by assuming that a given dataset involves a single classification task with $C$ classes. This enabled us to design $\textbf{y}\in{\rm I\!R}^{K}$ as a single categorical variable, $y \in \{1,...,C\}$, where $K=1$.\footnote{For multiple classification tasks (e.g., identity recognition and eye-glass detection tasks for face images), our method can be applied with multidimensional $\textbf{y}$, where each dimension, $y_k$, corresponds to one categorical variable for each task.} We represent $y$ as a $C$-dimensional one-hot vector. For $\textbf{z}\in{\rm I\!R}^{J}$, we assume the existence of multiple style factors and expect each factor to be captured by each variable, $z_j$, within $\textbf{z}$.

The training image dataset consists of labeled set $L = \{ (\textbf{x}^{(1)},t^{(1)}),...,(\textbf{x}^{(N_{L})},t^{(N_{L})}) \}$, where the $n$-th image, $\textbf{x}^{(n)}$, is paired with the corresponding class label, $t^{(n)} \in \{1,...,C\}$, and unlabeled set $U = \{ \textbf{x}^{(1)},...,\textbf{x}^{(N_{U})} \}$. Here, $N_{L}$ and $N_{U}$ are the numbers of samples in datasets $L$ and $U$, respectively, and $N_{L} \ll N_{U}$. The empirical data distributions over $L$ and $U$ are denoted by $q^L(\textbf{x},t)$ and $q^U(\textbf{x})$, respectively. 

\subsubsection{Semi-supervised Learning Objective}
To update encoder parameter $\phi$ and decoder parameter $\theta$, the objectives for the labeled and unlabeled sets are given as

\vskip -0.6cm

 \begin{equation} \label{eqn_loss_label_cat}
\mathcal{J}^L (\phi,\theta) = \mathcal{J}_{recon}^L (\phi,\theta)+\mathcal{J}_{cls}(\phi)-\mathcal{J}^{Both} (\phi)
\end{equation}

\vskip -0.6cm

 \begin{equation} \label{eqn_loss_unlabel_cat}
\mathcal{J}^U (\phi,\theta) = \mathcal{J}_{recon}^U (\phi,\theta)- \mathcal{J}^{Both} (\phi).
\end{equation}

\vskip -0.1cm

For the following reconstruction terms, we use true label $t$ for $L$ and inferred feature $y$ for $U$. This strategy helps the decoder accurately recognize one-hot class vectors.

\vskip -0.6cm

\begin{equation} \label{eqn_lossRecon_cat}
\mathcal{J}_{recon}^L (\phi,\theta) = \mathbb{E}_{q^L(\textbf{x},t)} \Big[ \mathbb{E}_{q_{\phi}(\textbf{z}|\textbf{x})} \big[ \log p_{\theta}(\textbf{x}|t,\textbf{z}) \big] \Big]
\end{equation}

\vskip -0.8cm

\begin{equation} \label{eqn_lossRecon}
\mathcal{J}_{recon}^U (\phi,\theta) = \mathbb{E}_{q^U(\textbf{x})} \Big[ 
\mathbb{E}_{q_{\phi}(y,\textbf{z}|\textbf{x})} \big[ 
\log p_{\theta}(\textbf{x}|y,\textbf{z}) \big] \Big]
\end{equation}

\vskip -0.2cm

We compute the classification term for $L$ as

\vskip -0.37cm

\begin{equation} \label{eqn_lossCls_cat}
\mathcal{J}_{cls} (\phi) = \alpha \, \mathbb{E}_{q^L(\textbf{x},t)} \Big[ 
\mathbb{I}(y=t) \log q_{\phi}(y|\textbf{x}) \Big],
\end{equation}

\vskip -0.13cm

\noindent where hyper-parameter $\alpha$ controls the effect of discriminative learning. With scaling constant $\rho$, we set $\alpha=\rho\frac{N_L+N_U}{N_L}$ \cite{nips14_kingmaDgm, icml16_adgm_auxiliary}.

Next, we introduce the following commonly used objective for $L$ and $U$ to learn disentangled features and regularize the encoder:

\vskip -0.6cm

\begin{equation} \label{eqn_lossBoth}
\begin{split}
\mathcal{J}^{Both} (\phi) =&\; \lambda \, \mathcal{L}^{y,z}_{VecIdp}(\phi) +\gamma_y \, \mathcal{L}^y_{Reg} (\phi) \\ & + \beta_z \, \mathcal{L}^{z}_{TC}(\phi) + \gamma_z \, \mathcal{L}^z_{Reg} (\phi). 
\end{split}
\end{equation}

\vskip -0.23cm

This is identical to applying individual loss weights to the vector independence, TC, and dimension-wise regularization terms in our KL decomposition (\ref{eqn_decompElbo}). Here, the expectation over the empirical distribution, i.e., $\mathbb{E}_{q^U(\textbf{x})} [ \cdot ]$ or $\mathbb{E}_{q^L(\textbf{x},t)} [ \cdot ]$, is included in computing the aggregated posteriors, $q_{\phi}($\textbf{z}$)$ and $q_{\phi}(y)$. Note that $ \mathcal{L}^{y}_{TC}$ disappears for a single class variable, $y$ (i.e., $K=1$ in (\ref{eqn_lossVar_sepa})), and data-latent MI $I_{q_{\phi}}(y,\textbf{z};\textbf{x})$ is removed so as to allow $y$ and $\textbf{z}$ to properly store the information about $\textbf{x}$.

The final optimization function is given as 

\vskip -0.4cm

\begin{equation} \label{eqn_loss_final}
\underset{\phi,\theta}{\text{maximize}} \; \mathcal{J}^L (\phi,\theta) +\mathcal{J}^U (\phi,\theta)
\end{equation}

\section{Data and Experimental Settings}

\subsection{Data and Experimental Settings}
We used the dSprites \cite{dsprites17}, Fashion-MNIST \cite{xiao2017fashionmnist}, and MNIST \cite{lecun2010mnist} datasets. For SSL, the labeled data were selected to be distributed evenly across classes, and the entire training set was used for the unlabeled set. We prevented overfitting to training data in classification tasks by introducing validation data. For the dSprites dataset, we divided the images in a ratio of 10:1:1 for training, validation, and testing and tested two SSL setups with 2\% and 0.25\% labeled training data. For the Fashion-MNIST and MNIST datasets, we divided the training set in a ratio of 5:1 for training and validation while maintaining the original test set and tested the SSL setup with 2\% labeled training data.

The architectures of the encoder and decoder were the same as the convolutional networks used in \cite{nips18_jointVae, icml19_cascadeVae}. The priors were set as $p(\textbf{z})=N(\textbf{0},\textbf{I})$ and $p(y)=Cat(\textbf{\textpi})$, where \textbf{\textpi} denotes evenly distributed class probabilities. We employed the Gumbel-Softmax distribution \cite{iclr17_gumbel, iclr17_concrete} for reparametrizing categorical $y$. We trained networks with minibatch weighted sampling \cite{nips18_betaTcVae}. Further details of experimental settings are described in Section S4.

We considered the vanilla VAE \cite{iclr14_vae}, $\beta$-VAE \cite{iclr17_betaVae}, $\beta$-TCVAE \cite{nips18_betaTcVae}, and jointVAE \cite{nips18_jointVae} as the baselines. For fair comparison, we augmented their original unsupervised objectives with the classification term in (\ref{eqn_lossCls_cat}) and the same loss weight, $\alpha$. For the VAE, $\beta$-VAE, and $\beta$-TCVAE, we augmented continuous vector $\textbf{z}$ in their original objectives with discrete variable $y$. Note that the main difference between the $\beta$-TCVAE and our IV-VAE is the existence of the vector independence term in (\ref{eqn_lossBoth}). We also removed the data-latent MI term from the $\beta$-TCVAE objective, as applied in our objective. See Section S5 for the baseline details.

\subsection{Performance Metrics}
\begin{figure*}[t]
\begin{center}
\centerline{\includegraphics[width=16.8cm]{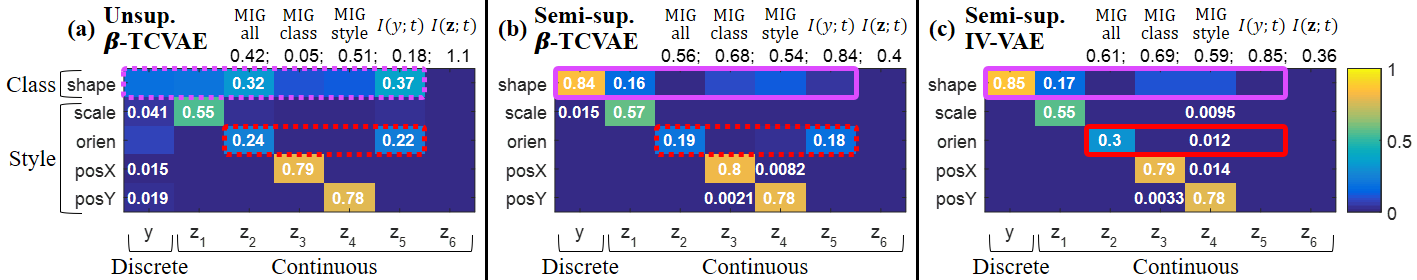}}
\vskip -0.2cm
\caption{Examples of disentanglement scores (top right) and empirical mutual information (bottom) between the ground truth generative factors of dSprites (y-axis) and inferred features (x-axis). (a) Unsupervised $\beta$-TCVAE ($\beta_z=4$). (b) Semi-supervised $\beta$-TCVAE ($\beta_z=4$). (c) Semi-supervised IV-VAE (ours; $\beta_z=4$ and $\lambda=4$). The three networks were initiated from the same random seed. The dotted-line boxes indicate undesirable results, while the solid-line boxes indicate improved results. The SSL with weak class supervision helps $y$ better capture discrete class factors (pink boxes). The vector independence objective helps $y$ and $\textbf{z}$ store different information, leading to better disentanglement (red boxes).}
\label{fig_score_re200310}
\end{center}
\vskip -0.5cm
\end{figure*}

We measured the classification error with $y$ to assess the amount of class information in $y$ and the ELBO to examine the generative modeling ability. As the disentanglement score, we computed the MI gap \cite{nips18_betaTcVae} based on the empirical MI between latent variables and known generative factors: $\text{MIG}=\frac{1}{M} \sum_{m=1}^{M} \text{MIG}_m $, where $M$ is the number of known factors and $\text{MIG}_m = \frac{1}{H(v_m)} \big(  I(h_{i^{(m)}};v_m) - \max_{i \neq i^{(m)}} I(h_i;v_m) \big)$ is the score for each factor, $v_m$, for quantifying the gap between the top two variables with the highest MI. Here, $I(h_i;v_m)/H(v_m)$ is the normalized MI between latent variable $h_i$ and factor $v_m$, and it is theoretically bounded between 0 and 1. Additionally, $i^{(m)}=\argmax_i I(h_i;v_m)$. This vanilla MIG is denoted by $\text{MIG}_{\text{all}}$. Figure \ref{fig_score_re200310} shows the example of the normalized MI computed on dSprites with different models, where the top two MI values for each factor are indicated.


\begin{figure*}[t]
\begin{center}
\centerline{\includegraphics[width=16.8cm]{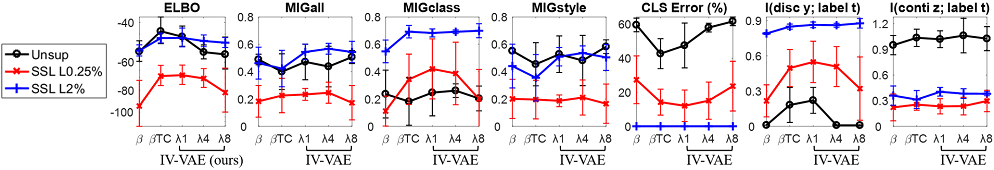}}
\vskip -0.2cm
\caption{Comparison on dSprites under the unsupervised (black) and semi-supervised setups with 0.25\% (red) and 2\% (blue) labeled data. For each network setting, the mean and std scores from 7 random seeds are shown. The x-axis shows $\beta$-VAE, $\beta$-TCVAE, and our IV-VAE trained with the same weight of ELBO KL or TC term ($\beta=\beta_z = 4$). The $\lambda$ weight for vector independence varies as 1, 4, and 8. Unsupervised learning causes the class factor to be stored in the continuous $\textbf{z}$ instead of the discrete $y$ (i.e., high $I(\textbf{z}; t)$ and low $I(y; t)$). Furthermore, the class information is captured in multiple variables, causing lower $\text{MIG}_{\text{class}}$ than $\text{MIG}_{\text{style}}$. Introducing weak supervision with a few labeled data effectively relieves this issue (i.e., low $I(\textbf{z}; t)$ and high $I(y; t)$). Our vector independence objective allows $y$ and $\textbf{z}$ to capture different information, improving most of the scores. Keep in mind that the baselines were also trained under the same SSL setup using $\textbf{z}$ and the discrete $y$.}
\label{fig_shp_metric_200206}
\end{center}
\vskip -0.6cm
\end{figure*}

The entangled results for the case where one variable captures both style and class factors (e.g., $z_2$ in Figure \ref{fig_score_re200310}(a)) can obtain fairly good $\text{MIG}_{\text{all}}$ scores. To alleviate this issue, we \textit{separately} computed the MIG for the class and style factors as $\text{MIG}_{\text{class}}=\text{MIG}_t$ (i.e., the MIG of the first row in Figure \ref{fig_score_re200310}) and $\text{MIG}_{\text{style}}=\frac{1}{|S|} \sum_{m \in S} \text{MIG}_m$ (i.e., the mean MIG averaged from the second to last rows), where $t$ denotes ground truth class labels and $S$ is the set of style factors. For the cases \textit{without} known generative factors but \textit{with} available labels, $t$, we also computed the MI that assesses how much $y$ represents the class information and $\textbf{z}$ does not as $I(y; t)$ and $I(\textbf{z}; t)=\sum_{j} I(z_j; t)$, respectively.

\section{Experimental Results}

All results presented in this section were obtained using test data. See our supplementary material for additional results.

\subsection{Results on dSprites}

\subsubsection{Effect of SSL and Vector Independence}
Figure \ref{fig_score_re200310} depicts the benefit of SSL and vector independence. Using purely unsupervised learning failed to capture the class factor with discrete variable $y$, but employing SSL with 2\% class labels easily resolved this issue. Encouraging vector independence helped $\textbf{z}$ better capture the style factors by forcing $y$ and $\textbf{z}$ to store different information.

Figure \ref{fig_shp_metric_200206} shows the results on dSprites obtained under various SSL setups. The unsupervised setup yielded good vanilla $\text{MIG}_{\text{all}}$ scores, which were similar to those of the 2\%-labeled setup. However, the unsupervised setup caused the continuous vector $\textbf{z}$ to mostly captured the class information and the discrete $y$ to fail to store it. This is evidenced by higher $I(\textbf{z}; t)$ and lower $I(y; t)$ than those of the SSL setups. Furthermore, as indicated by lower $\text{MIG}_{\text{class}}$ scores, more than two variables undesirably captured the class factor under the unsupervised setup.

Injecting weak class supervision to the training process simply yet effectively alleviated this difficulty, as shown in the decreased $I(\textbf{z};t)$ and increased $I(y; t)$ in Figure \ref{fig_shp_metric_200206}. In addition, our IV-VAEs with proper $\lambda$ weights outperformed $\beta$-VAEs and $\beta$-TCVAEs for most scores. In particular, the score improvements with the 0.25\% labels were larger than those with the 2\% labels, indicating the benefit of vector independence under a few class labels. In terms of the ELBO, our IV-VAEs did not outperform $\beta$-TCVAEs but showed less trade-off between density modeling and disentanglement than $\beta$-VAE (i.e., higher ELBO and MIG scores).


\begin{figure*}[]
\begin{center}
\centerline{\includegraphics[width=16.8cm]{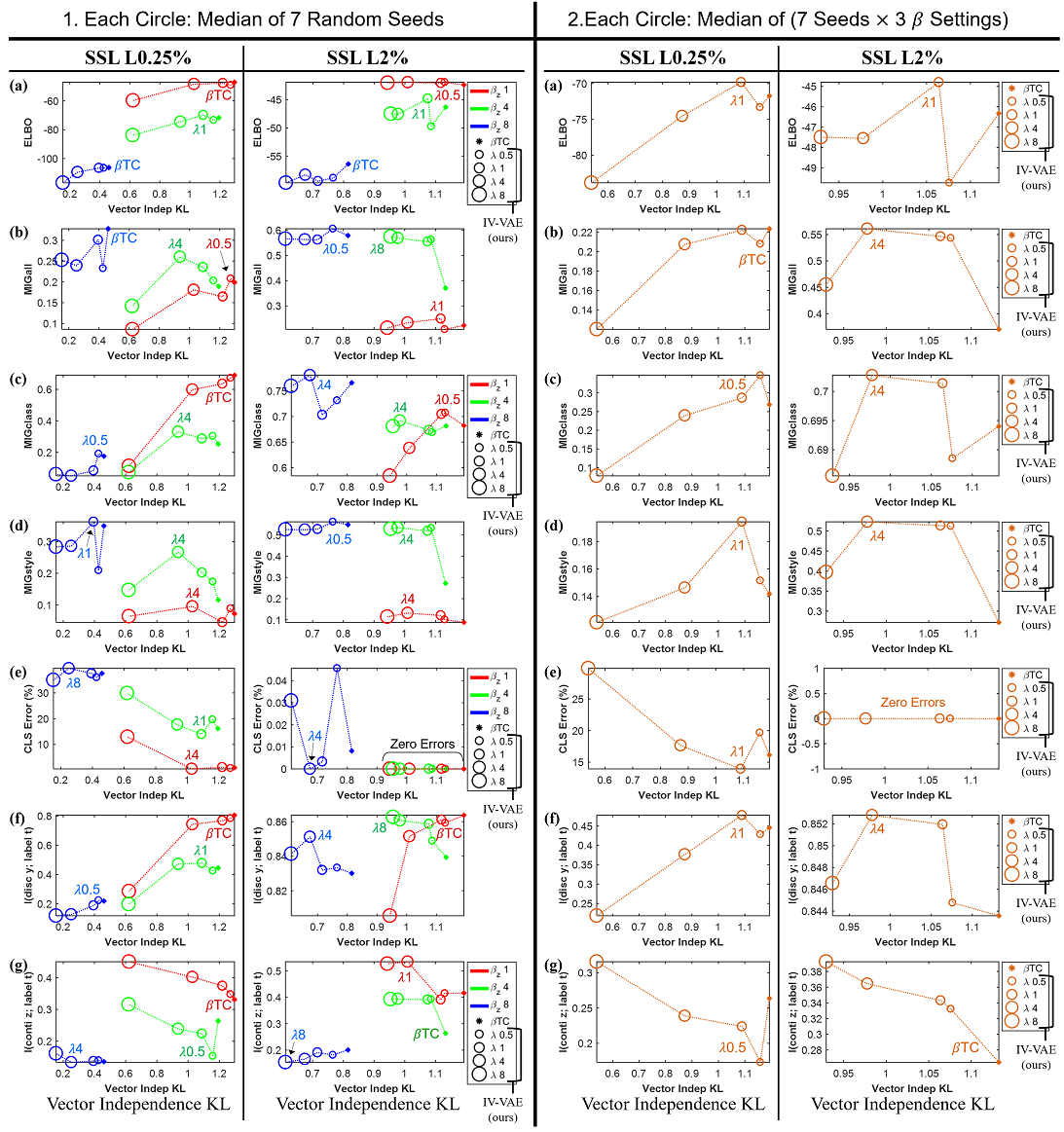}}
\caption{Relationship between the vector independence objective and the evaluation metrics. The x-axis shows vector independence KL $\mathcal{L}^{y,z}_{VecIdp} = \textrm{D}_{\textrm{KL}}\big( q_{\phi}(y,\textbf{z}) || q_{\phi}(y)q_{\phi}(\textbf{z})\big)$, where lower KL values indicate stronger independence. The y-axis shows (a) ELBO, (b) $\text{MIG}_{\text{all}}$, (c) $\text{MIG}_{\text{class}}$, (d) $\text{MIG}_{\text{style}}$, (e) classification error, (f) $I(y; t)$, and (g)$I(\textbf{z}; t)$ obtained under the SSL setups with 0.25\% and 2\% labeled data (denoted by SSL L0.25\% and SSL L2\%, respectively). Lower is better for (e) and (g), and higher is better for the other metrics. In Figure \ref{fig_shp_relation_200207}-1 (left), each circle shows the median score of seven networks initiated from different random seeds. As shown in the legend, different colors indicate different $\beta_z$ values for controlling the effect of TC in training, and bigger circles indicate bigger $\lambda$ values for causing stronger vector independence. In Figure \ref{fig_shp_relation_200207}-2 (right), the different $\beta_z$ settings are further merged to analyze the overall tendency, and each circle shows the median score of 21 networks (i.e., 7 random seeds $\times$ 3 $\beta_z$ settings). Given each setting, the best performing network is indicated with text. As shown in Figure \ref{fig_shp_relation_200207}-1, given the value of $\beta_z$, IV-VAEs often outperform $\beta$-TCVAEs. As shown in Figure \ref{fig_shp_relation_200207}-2, IV-VAEs with the $\lambda$ of 1 for the 0.25\% SSL setup and those with the $\lambda$ of 4 for the 2\% SSL setup generally work well.}
\label{fig_shp_relation_200207}
\end{center}
\end{figure*}


\subsubsection{Relationships between Vector Independence and Evaluation Metrics}
Figure \ref{fig_shp_relation_200207} shows the relationships between the vector independence objective and evaluation metrics. In Figure \ref{fig_shp_relation_200207}-1 (left), we depicted scatter plots, where each circle represents the median score of seven networks trained with the same loss weights but initiated from different random seeds. In Figure \ref{fig_shp_relation_200207}-2 (right), the different $\beta_z$ values were further integrated to analyze the general result trends. Notice that in the figures, a larger $\lambda$  led to a smaller KL of vector independence, i.e., a stronger independence between $y$ and $\textbf{z}$. The analyses are summarized below. See Section S6 for the additional analyses with extended baselines.

\vskip -2cm

\begin{itemize}[noitemsep]

\setlength{\leftskip}{-0.3cm}

\item \textbf{ELBO} in Figure \ref{fig_shp_relation_200207}-1(a). The ELBO values of IV-VAEs were similar to those of $\beta$-TCVAEs for lower $\lambda$ values of 0.5 and 1. Increasing $\lambda$ caused slightly decreased ELBO values, because the models focused more on learning disentangled features than maximizing the reconstruction term. Nevertheless, under higher $\beta_z$ values of 4 and 8, IV-VAEs yielded better ELBO than $\beta$-VAEs\footnote{The $\beta$-VAEs with $\beta$ of 4 and 8 under the 0.25\% SSL setup yielded the median ELBO of -102.2 and -120.8, respectively. Those with $\beta$ of 4 and 8 under the 2\% SSL setup yielded the median ELBO of -53.9 and -72.5.} because we did not penalize the latent-data MI term, which led to an eased trade-off between reconstruction and disentanglement.

\vskip 0.15cm

\item \textbf{$\text{MIG}_{\text{all}}$, $\text{MIG}_{\text{class}}$, and $\text{MIG}_{\text{style}}$} in Figure \ref{fig_shp_relation_200207}-1(b), (c), and (d). For most of the $\beta_z$ setups, IV-VAEs with the $\lambda$ values of 0.5, 1, or 4 achieved better MIG scores than $\beta$-TCVAEs, showing that reducing the vector independence objective along with the TC helps disentanglement. Higher $\beta_z$ that heavily penalized the TC often led to higher MIG scores (i.e., the blue and green lines showed better scores than the red lines). The optimal $\lambda$ yielding the highest MIG differed depending on the value of $\beta_z$, but the $\lambda$ of 0.5, 1, and 4 generally worked well.

\vskip 0.15cm

\item \textbf{Classification error and $\textbf{I(y; t)}$} in  Figure \ref{fig_shp_relation_200207}-1(e) and (f). Given the value of $\beta_z$, the lowest classification error and highest $I(y; t)$ score were often obtained with IV-VAE, in comparison to $\beta$-TCVAE. This result implies that vector independence encourages $y$ to better capture the class factor. In the SSL setup with 0.25\% labels, excessively increasing $\lambda$ values harmed the classification performance. 

\vskip 0.15cm

\item \textbf{$\textbf{I(\textbf{z}; t)}$} in Figure \ref{fig_shp_relation_200207}-1(g). IV-VAEs often achieved lower $I(\textbf{z}; t)$ than that of $\beta$-TCVAEs. This result implies that vector independence prevents $\textbf{z}$ from storing the class factor by enforcing $\textbf{z}$ and $y$ to represent different information, leading to better disentanglement.

\vskip 0.15cm

\item \textbf{Overall result trends} in Figure \ref{fig_shp_relation_200207}-2. Our IV-VAEs outperformed $\beta$-TCVAEs for most scores, demonstrating the benefit of vector independence. In general, the $\lambda$ of 1 for the 0.25\% setup and the $\lambda$ of 4 for the 2\% setup worked well.  

\setlength{\leftskip}{0pt}

\end{itemize}

\begin{figure}[t]

\begin{center}
\centerline{\includegraphics[width=\columnwidth]{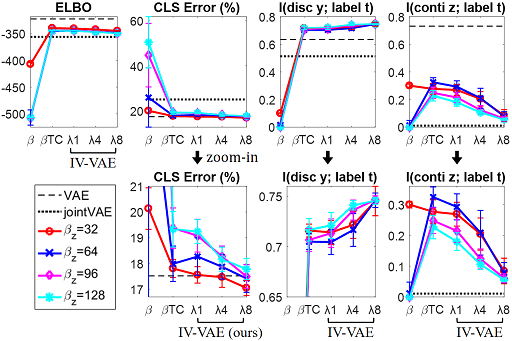}}
\vskip -0.2cm
\caption{Comparison on Fashion-MNIST with 2\% labeled data. For each network setting, the mean and std scores from 10 random seeds are shown. The horizontal lines indicate the mean performance of vanilla VAE and jointVAE. The x-axis shows $\beta$-VAE, $\beta$-TCVAE, and our IV-VAEs.  Our IV-VAEs outperform the baselines for most scores, demonstrating the benefit of vector independence. Keep in mind that the baselines were also trained under the same SSL setup using $\textbf{z}$ and the discrete $y$.
}
\label{fig_fashMni_result_2002006}
\end{center}
\vskip -0.5cm
\end{figure}


\begin{figure}[t]
\begin{center}
\centerline{\includegraphics[width=\columnwidth]{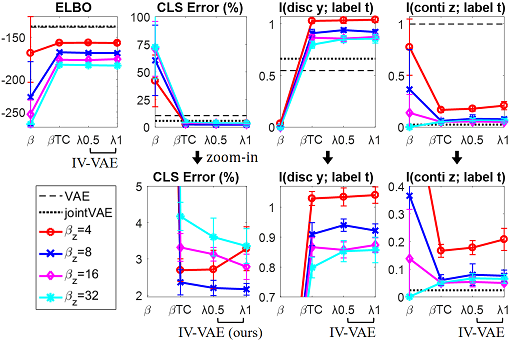}}
\vskip -0.2cm
\caption{Comparison on MNIST with 2\% labeled data. The settings and results are similar to those of Figure \ref{fig_fashMni_result_2002006}.
}
\label{fig_mnist_result_2002006}
\end{center}
\vskip -0.6cm
\end{figure}


\subsection{Results on Fashion-MNIST and MNIST}

Figure \ref{fig_fashMni_result_2002006} shows the results on Fashion-MNIST with 2\% labeled data. Because of the absence of ground truth style factors, we only measured the MI between the latent variables and class labels. For most values of $\beta_{z}$, promoting vector independence with greater $\lambda$ allowed $y$ and $\textbf{z}$ to capture different factors, causing more class information to be stored in $y$. Our IV-VAEs achieved better classification errors and $I(y; t)$ as well as lower $I(\textbf{z}; t)$ than all the baselines, indicating improved disentanglement. Figure \ref{fig_mnist_result_2002006} shows the results on MNIST with 2\% labeled data. We also observed the benefit of our vector independence under most of the $\beta$ weight settings.

\begin{figure}[]

\begin{center}
\centerline{\includegraphics[width=\columnwidth]{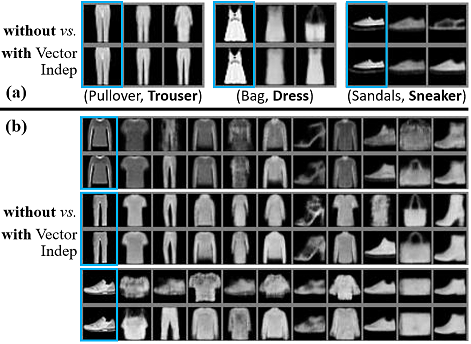}}
\caption{Qualitative results on Fashion-MNIST. (a) Corrected classification. Given an input (left), inferred class probabilities (middle) or one-hot labels (right) from the class encoding $y$ are used for reconstruction. Below each example, inferred labels without and with the vector independence loss are indicated. (b) Style-conditional generation. The style vector $\textbf{z}$ is extracted from an input (leftmost), and the $y$ is set as desired item labels (right). Without promoting vector independence, the network often fails to generate Trouser (the 2nd column in the generation results), Sneaker (the 8th), and Bag (the 9th) classes. Encouraging vector independence forces $y$ to better capture class information in (a) and suitably separates styles from clothing classes in (b).}
\label{fig_fashMni_Ex200206}
\end{center}
\vskip -1.1cm
\end{figure}

\begin{figure}[]

\begin{center}
\centerline{\includegraphics[width=\columnwidth]{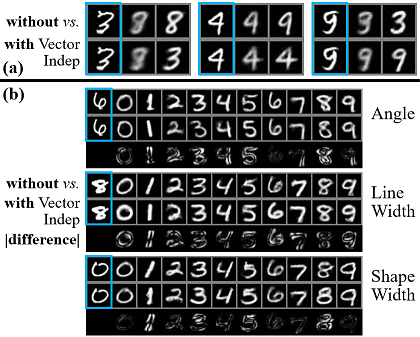}}
\caption{Qualitative results on MNIST. The analysis is similar to that of Figure \ref{fig_fashMni_Ex200206}. (a) Corrected classification. (b) Style-conditional generation. In (b), the absolute differences of pixel values are also shown. Encouraging vector independence better exhibits the digit styles (e.g., writing angle, line width, and shape width) by helping $\textbf{z}$ effectively capture style information.}
\label{fig_mnist_Ex200206}
\end{center}
\end{figure}


Figures \ref{fig_fashMni_Ex200206} and \ref{fig_mnist_Ex200206} depict the qualitative results of the two networks trained without and with the vector independence objective. The networks were obtained from the same random seed and loss weights except the $\lambda$ weight.\footnote{In Figure \ref{fig_fashMni_Ex200206}, the $\lambda$ of 4 was used for training the IV-VAE, while the $\beta_z$ of 96, $\gamma_z$ of 1, and $\gamma_y$ of 2 were commonly used for both networks. In Figure \ref{fig_mnist_Ex200206}, the $\lambda$ of 8 was used, while the $\beta_z$ of 32, $\gamma_z$ of 1, and $\gamma_y$ of 2 were the same for both networks.} See Section S7 and S8 for additional examples.

To visually show the corrected class labels by encouraging vector independence, Figures \ref{fig_fashMni_Ex200206}(a) and \ref{fig_mnist_Ex200206}(a) depict reconstruction examples with estimated class information, i.e., inferred class probabilities or one-hot labels via the argmax operation. The results obtained from the class probabilities were often blurry, indicating that the inputs were confusing to be classified. Employing vector independence often corrects the classification results by enforcing $y$ to better store the class factor.

Figures \ref{fig_fashMni_Ex200206}(b) and \ref{fig_mnist_Ex200206}(b) depict the generation examples given input styles. We extracted $\textbf{z}$ from the input and set $y$ as each of the one-hot class labels. Enhancing vector independence better displays the fashion and digit styles by helping $\textbf{z}$ effectively capture style information.


\section{Conclusion} \label{sect_conclusion}

We have proposed an approach for semi-supervised disentanglement learning with a variational autoencoder. In our method, two latent vectors separately capture class and style factors. To boost disentanglement, we have proposed the vector independence objective that enforces the vectors to be statistically independent. We have revealed that, along with the total correlation term, our vector independence term is another source of disentanglement in the evidence lower bound. Furthermore, the difficulty in the learning of discrete factors can be reduced by exploiting a small number of class labels. The experiments on the dSprites, Fashion-MNIST, and MNIST datasets have confirmed the effectiveness of our method for image classification and generation.

\section*{Acknowledgement}
We thank Prof. Sung Ju Hwang (KAIST) for providing his valuable comments. This work was supported by an Institute for Information and Communications Technology Promotion (IITP) grant funded by the Korean Government (MSIT) under grant no. 2016-0-00562 (R0124-16-0002).

\bibliography{bib_ivvae_arxiv.bib}

\begin{thebibliography}{28}
\providecommand{\natexlab}[1]{#1}
\providecommand{\url}[1]{\texttt{#1}}
\expandafter\ifx\csname urlstyle\endcsname\relax
  \providecommand{\doi}[1]{doi: #1}\else
  \providecommand{\doi}{doi: \begingroup \urlstyle{rm}\Url}\fi

\bibitem[Amari et~al.(1996)Amari, Cichocki, and Yang]{nips96_ica_2_amari}
Amari, S.-i., Cichocki, A., and Yang, H.~H.
\newblock A new learning algorithm for blind signal separation.
\newblock In \emph{Proc. NIPS}, pp.\  757--763, 1996.

\bibitem[Bengio et~al.(2013)Bengio, Courville, and
  Vincent]{tpami13_bengio_reprLearn}
Bengio, Y., Courville, A., and Vincent, P.
\newblock Representation learning: A review and new perspectives.
\newblock \emph{IEEE Trans. Pattern Anal. Mach. Intell.}, 35\penalty0
  (8):\penalty0 1798--1828, 2013.

\bibitem[Burgess et~al.(2017)Burgess, Higgins, Pal, Matthey, Watters,
  Desjardins, and Lerchner]{nipsw17_bVae_anneal}
Burgess, C.~P., Higgins, I., Pal, A., Matthey, L., Watters, N., Desjardins, G.,
  and Lerchner, A.
\newblock Understanding disentangling in beta-vae.
\newblock In \emph{Proc. NIPS Workshop}, 2017.

\bibitem[Chen et~al.(2018)Chen, Li, Grosse, and Duvenaud]{nips18_betaTcVae}
Chen, T.~Q., Li, X., Grosse, R.~B., and Duvenaud, D.~K.
\newblock Isolating sources of disentanglement in variational autoencoders.
\newblock In \emph{Proc. NIPS}, pp.\  2610--2620, 2018.

\bibitem[Dupont(2018)]{nips18_jointVae}
Dupont, E.
\newblock Learning disentangled joint continuous and discrete representations.
\newblock In \emph{Proc. NIPS}, pp.\  710--720, 2018.

\bibitem[Esmaeili et~al.(2019)Esmaeili, Wu, Jain, Bozkurt, Siddharth, Paige,
  Brooks, Dy, and van~de Meent]{aistats19_structDisent}
Esmaeili, B., Wu, H., Jain, S., Bozkurt, A., Siddharth, N., Paige, B., Brooks,
  D.~H., Dy, J., and van~de Meent, J.-W.
\newblock Structured disentangled representations.
\newblock In \emph{Proc. AISTATS}, pp.\  2525--2534, 2019.

\bibitem[Gao et~al.(2019)Gao, Brekelmans, Steeg, and
  Galstyan]{aistats19_gaoAutoEncTc}
Gao, S., Brekelmans, R., Steeg, G.~V., and Galstyan, A.
\newblock Auto-encoding total correlation explanation.
\newblock In \emph{Proc. AISTATS}, pp.\  1157--1166, 2019.

\bibitem[Higgins et~al.(2017)Higgins, Matthey, Pal, Burgess, Glorot, Botvinick,
  Mohamed, and Lerchner]{iclr17_betaVae}
Higgins, I., Matthey, L., Pal, A., Burgess, C., Glorot, X., Botvinick, M.,
  Mohamed, S., and Lerchner, A.
\newblock beta-vae: Learning basic visual concepts with a constrained
  variational framework.
\newblock In \emph{Proc. ICLR}, 2017.

\bibitem[Hoffman \& Johnson(2016)Hoffman and Johnson]{nipsw16_elboSurgery}
Hoffman, M.~D. and Johnson, M.~J.
\newblock Elbo surgery: yet another way to carve up the variational evidence
  lower bound.
\newblock In \emph{Proc. NIPS Workshop}, 2016.

\bibitem[Jang et~al.(2017)Jang, Gu, and Poole]{iclr17_gumbel}
Jang, E., Gu, S., and Poole, B.
\newblock Categorical reparameterization with gumbel-softmax.
\newblock In \emph{Proc. ICLR}, 2017.

\bibitem[Jeong \& Song(2019)Jeong and Song]{icml19_cascadeVae}
Jeong, Y. and Song, H.~O.
\newblock Learning discrete and continuous factors of data via alternating
  disentanglement.
\newblock In \emph{Proc. ICML}, pp.\  3091--3099, 2019.

\bibitem[Jutten \& Herault(1991)Jutten and Herault]{sp91_ica_1}
Jutten, C. and Herault, J.
\newblock Blind separation of sources, part i: An adaptive algorithm based on
  neuromimetic architecture.
\newblock \emph{Signal Processing}, 24\penalty0 (1):\penalty0 1--10, 1991.

\bibitem[Kim \& Mnih(2018)Kim and Mnih]{icml18_factorVae}
Kim, H. and Mnih, A.
\newblock Disentangling by factorising.
\newblock In \emph{Proc. ICML}, pp.\  2649--2658, 2018.

\bibitem[Kim et~al.(2006{\natexlab{a}})Kim, Attias, Lee, and Lee]{iva_2_jour}
Kim, T., Attias, H.~T., Lee, S.-Y., and Lee, T.-W.
\newblock Blind source separation exploiting higher-order frequency
  dependencies.
\newblock \emph{IEEE Trans. Audio, Speech, Language Process.}, 15\penalty0
  (1):\penalty0 70--79, 2006{\natexlab{a}}.

\bibitem[Kim et~al.(2006{\natexlab{b}})Kim, Eltoft, and Lee]{iva_1_conf}
Kim, T., Eltoft, T., and Lee, T.-W.
\newblock Independent vector analysis: An extension of ica to multivariate
  components.
\newblock In \emph{Proc. Int. Conf. on ICA}, pp.\  165--172,
  2006{\natexlab{b}}.

\bibitem[Kingma \& Welling(2014)Kingma and Welling]{iclr14_vae}
Kingma, D.~P. and Welling, M.
\newblock Auto-encoding variational bayes.
\newblock In \emph{Proc. ICLR}, 2014.

\bibitem[Kingma et~al.(2014)Kingma, Mohamed, Rezende, and
  Welling]{nips14_kingmaDgm}
Kingma, D.~P., Mohamed, S., Rezende, D.~J., and Welling, M.
\newblock Semi-supervised learning with deep generative models.
\newblock In \emph{Proc. NIPS}, pp.\  3581--3589, 2014.

\bibitem[Kumar et~al.(2018)Kumar, Sattigeri, and Balakrishnan]{iclr18_dipVae}
Kumar, A., Sattigeri, P., and Balakrishnan, A.
\newblock Variational inference of disentangled latent concepts from unlabeled
  observations.
\newblock In \emph{Proc. ICLR}, 2018.

\bibitem[LeCun et~al.(2010)LeCun, Cortes, and Burges]{lecun2010mnist}
LeCun, Y., Cortes, C., and Burges, C.
\newblock Mnist handwritten digit database.
\newblock http://yann.lecun.com/exdb/mnist/, 2010.

\bibitem[Locatello et~al.(2019)Locatello, Bauer, Lucic, Raetsch, Gelly,
  Sch{\"o}lkopf, and Bachem]{icml18_commAssump}
Locatello, F., Bauer, S., Lucic, M., Raetsch, G., Gelly, S., Sch{\"o}lkopf, B.,
  and Bachem, O.
\newblock Challenging common assumptions in the unsupervised learning of
  disentangled representations.
\newblock In \emph{Proc. ICML}, pp.\  4114--4124, 2019.

\bibitem[Maaløe et~al.(2016)Maaløe, Sønderby, Sønderby, and
  Winther]{icml16_adgm_auxiliary}
Maaløe, L., Sønderby, C.~K., Sønderby, S.~K., and Winther, O.
\newblock Auxiliary deep generative models.
\newblock In \emph{Proc. ICML}, pp.\  1445--1453, 2016.

\bibitem[Maddison et~al.(2017)Maddison, Mnih, and Teh]{iclr17_concrete}
Maddison, C.~J., Mnih, A., and Teh, Y.~W.
\newblock The concrete distribution: A continuous relaxation of discrete random
  variables.
\newblock In \emph{Proc. ICLR}, 2017.

\bibitem[Makhzani et~al.(2016)Makhzani, Shlens, Jaitly, Goodfellow, and
  Frey]{iclrw16_aae}
Makhzani, A., Shlens, J., Jaitly, N., Goodfellow, I., and Frey, B.
\newblock Adversarial autoencoders.
\newblock In \emph{Proc. ICLR Workshop}, 2016.

\bibitem[Matthey et~al.(2017)Matthey, Higgins, Hassabis, and
  Lerchner]{dsprites17}
Matthey, L., Higgins, I., Hassabis, D., and Lerchner, A.
\newblock dsprites: Disentanglement testing sprites dataset.
\newblock https://github.com/deepmind/dsprites-dataset/, 2017.

\bibitem[Narayanaswamy et~al.(2017)Narayanaswamy, Paige, Van~de Meent,
  Desmaison, Goodman, Kohli, Wood, and Torr]{nips17_disent_semiSupDgm}
Narayanaswamy, S., Paige, T.~B., Van~de Meent, J.-W., Desmaison, A., Goodman,
  N., Kohli, P., Wood, F., and Torr, P.
\newblock Learning disentangled representations with semi-supervised deep
  generative models.
\newblock In \emph{Proc. NIPS}, pp.\  5925--5935, 2017.

\bibitem[Watanabe(1960)]{ibm1960_tc}
Watanabe, S.
\newblock Information theoretical analysis of multivariate correlation.
\newblock \emph{IBM Journal of research and development}, 4\penalty0
  (1):\penalty0 66--82, 1960.

\bibitem[Xiao et~al.(2017)Xiao, Rasul, and Vollgraf]{xiao2017fashionmnist}
Xiao, H., Rasul, K., and Vollgraf, R.
\newblock Fashion-mnist: a novel image dataset for benchmarking machine
  learning algorithms.
\newblock In \emph{arXiv preprint arXiv:1708.07747}, 2017.

\bibitem[Zheng \& Sun(2019)Zheng and Sun]{cvpr19_ssl_genEx}
Zheng, Z. and Sun, L.
\newblock Disentangling latent space for vae by label relevant/irrelevant
  dimensions.
\newblock In \emph{Proc. CVPR}, pp.\  12192--12201, 2019.

\end{thebibliography}
\bibliographystyle{icml2020}


\onecolumn

\icmltitle{Supplementary Information for “Semi-supervised Disentanglement\\with Independent Vector Variational Autoencoders”}

\setcounter{section}{0}
\setcounter{figure}{0}
\renewcommand{\thefigure}{S\arabic{figure}}
\renewcommand{\thesection}{S\arabic{section}}

\icmltitlerunning{Supplementary Information for “Semi-supervised Disentanglement with Independent Vector Variational Autoencoders”}


\section*{Contents} \label{supple_toc}

\begin{enumerate}[noitemsep, label={S\arabic*.}]

\setlength{\leftskip}{1cm}

\item \hyperlink{supple_s1}{Variational Bound with Two Latent Vectors}
\item \hyperlink{supple_s2}{Our ELBO Decomposition}
\item \hyperlink{supple_s3_rel}{Relationships between Vector Independence and TC}
\item \hyperlink{supple_s4_expSetting}{Data and Experimental Settings}
\item \hyperlink{supple_s5_baselines}{Baseline Methods}

\item \hyperlink{supple_s6_dSprites}{In-depth Analysis of Relationships between Vector Independence and Evaluation Metrics}

\item \hyperlink{supple_s7_fashMni}{Additional Results on Fashion-MNIST}
\item \hyperlink{supple_s8_mnist}{Additional Results on MNIST}

\setlength{\leftskip}{0pt}
 \end{enumerate}

\section*{List of Figures} \label{supple_toc_fig}

\begin{enumerate}[noitemsep, label={S\arabic*.}]

\setlength{\leftskip}{1cm}

\item \hyperlink{link_fig_derive_vecIdp}{(Derivation) Our decomposition of the KL term in the variational bound with two latent vectors}

\item \hyperlink{link_fig_derive_rel_vecIdp_tc}{(Derivation) Relationship between the vector independence objective and the separate TC}

\item \hyperlink{link_fig_archi}{Network architecture}

\item \hyperlink{link_fig_metricRel_allTC}{Comparison of the collective TC and the separate TC}

\item \hyperlink{link_fig_fashMni_supple_corrEx}{Corrected classification examples on Fashion-MNIST}
\item \hyperlink{link_fig_fashMni_supple_genY_1clo}{Style-controlled generation on Fashion-MNIST}
\item \hyperlink{link_fig_fashMni_supple_genY_2item}{Style-controlled generation on Fashion-MNIST: continued from Figure \hyperlink{link_fig_fashMni_supple_genY_1clo}{S6}}
\item \hyperlink{link_fig_fashMni_supple_zTrav_1shape}{Latent traversal examples on Fashion-MNIST}
\item \hyperlink{link_fig_fashMni_supple_zTrav_2bright}{Latent traversal examples on Fashion-MNIST: continued from Figure \hyperlink{link_fig_fashMni_supple_zTrav_1shape}{S8}
}

\item \hyperlink{link_fig_digitMni_supple_corrEx}{Corrected classification examples on MNIST}
\item \hyperlink{link_fig_digitMni_supple_genY}{Style-controlled generation on MNIST}
\item \hyperlink{link_fig_digitMni_supple_zTrav}{Latent traversal examples on MNIST}

\setlength{\leftskip}{0pt}
 \end{enumerate}

\clearpage

\hypertarget{supple_s1}{\section{Variational Bound with Two Latent Vectors} }

Suppose that the generation process of the data involves two latent vectors, \textbf{y} and \textbf{z}, with a joint prior, $p(\textbf{y},\textbf{z})$. By introducing an approximate posterior, $q_{\phi}(\textbf{y},\textbf{z}|\textbf{x})$, and a likelihood, $p_{\theta}(\textbf{x}|\textbf{y},\textbf{z})$, the ELBO for a single data sample becomes

\vskip -0.4cm

\begin{equation} \label{eqn_elbo_unsup_x}
\begin{split}
\log p_{\theta}(\textbf{x}) &=\log \iint_{\textbf{y},\textbf{z}} p_{\theta}(\textbf{x},\textbf{y},\textbf{z})\, d\textbf{y}d\textbf{z}=\log \iint_{\textbf{y},\textbf{z}} \frac{ q_{\phi}(\textbf{y},\textbf{z}|\textbf{x}) }{  q_{\phi}(\textbf{y},\textbf{z}|\textbf{x})} p_{\theta}(\textbf{x},\textbf{y},\textbf{z})\, d\textbf{y}d\textbf{z} = \log \Big( \mathbb{E}_{q_{\phi}(\textbf{y},\textbf{z}|\textbf{x})} \Big[ \frac{ p_{\theta}(\textbf{x},\textbf{y},\textbf{z}) }{  q_{\phi}(\textbf{y},\textbf{z}|\textbf{x})} \Big] \Big)  \\ &  \geq \mathbb{E}_{q_{\phi}(\textbf{y},\textbf{z}|\textbf{x})} \Big[ \log\frac{ p_{\theta}(\textbf{x},\textbf{y},\textbf{z}) }{  q_{\phi}(\textbf{y},\textbf{z}|\textbf{x})} \Big]  \qquad \because \mathrm{Jensen's \ inequality} \\ &= \mathbb{E}_{q_{\phi}(\textbf{y},\textbf{z}|\textbf{x})} \Big[ \log\frac{ p_{\theta}(\textbf{x}|\textbf{y},\textbf{z}) p(\textbf{y},\textbf{z}) }{  q_{\phi}(\textbf{y},\textbf{z}|\textbf{x})} \Big] = \mathbb{E}_{q_{\phi}(\textbf{y},\textbf{z}|\textbf{x})} \Big[ \log  p_{\theta}(\textbf{x}|\textbf{y},\textbf{z}) \Big] -
\textrm{D}_{\textrm{KL}} \big( q_{\phi}(\textbf{y},\textbf{z} | \textbf{x}) || p(\textbf{y},\textbf{z}) \big).
\end{split}
\end{equation}

\hypertarget{supple_s2}{\section{Our ELBO Decomposition} }

Figure \hyperlink{link_fig_derive_vecIdp}{S1} verifies our decomposition of the KL term, $ \mathbb{E}_{q(\textbf{x})} \big[ \textrm{D}_{\textrm{KL}} \big( q_{\phi}(\textbf{y},\textbf{z} | \textbf{x}) || p(\textbf{y},\textbf{z}) \big) \big] $, in the ELBO on $\mathbb{E}_{q(\textbf{x})} \big[\log p_{\theta}(\textbf{x}) \big]$.


\hypertarget{supple_s3_rel}
{\section{Relationships between Vector Independence and TC} }

Figure \hyperlink{link_fig_derive_rel_vecIdp_tc}{S2} verifies the following relationship: perfect vector independence ensures that the collective TC becomes the sum of the two separate TCs (i.e., $\mathcal{L}^{y,z}_{VecIdp}=0 \; \Rightarrow \; \mathcal{L}^{y,z}_{TC}=\mathcal{L}^{y}_{TC}+\mathcal{L}^{z}_{TC}$).

Moreover, in this document, we provide the experimental results regarding the relationship described in (14) of the main paper: $\mathcal{L}^{y,z}_{TC}=0 \; \Rightarrow  \; \mathcal{L}^{y}_{TC}=\mathcal{L}^{z}_{TC}=\mathcal{L}^{y,z}_{VecIdp}=0$. We show that (i) a perfectly penalized collective TC (i.e., $\mathcal{L}^{y,z}_{TC}=0$) rarely occurs because of the existence of other loss terms (e.g., a reconstruction term) and (ii) encouraging vector independence under penalizing either the separate TC $\mathcal{L}^{z}_{TC}$ or the collective TC $\mathcal{L}^{y,z}_{TC}$ can improve disentanglement performance. See Section \hyperlink{supple_s5_baselines}{S5} for the experimental settings and Section \hyperlink{supple_s6_dSprites}{S6} for the results.

\vskip -0.2cm

\begin{figure}[]
\begin{center}
\centerline{\includegraphics[width=17cm]{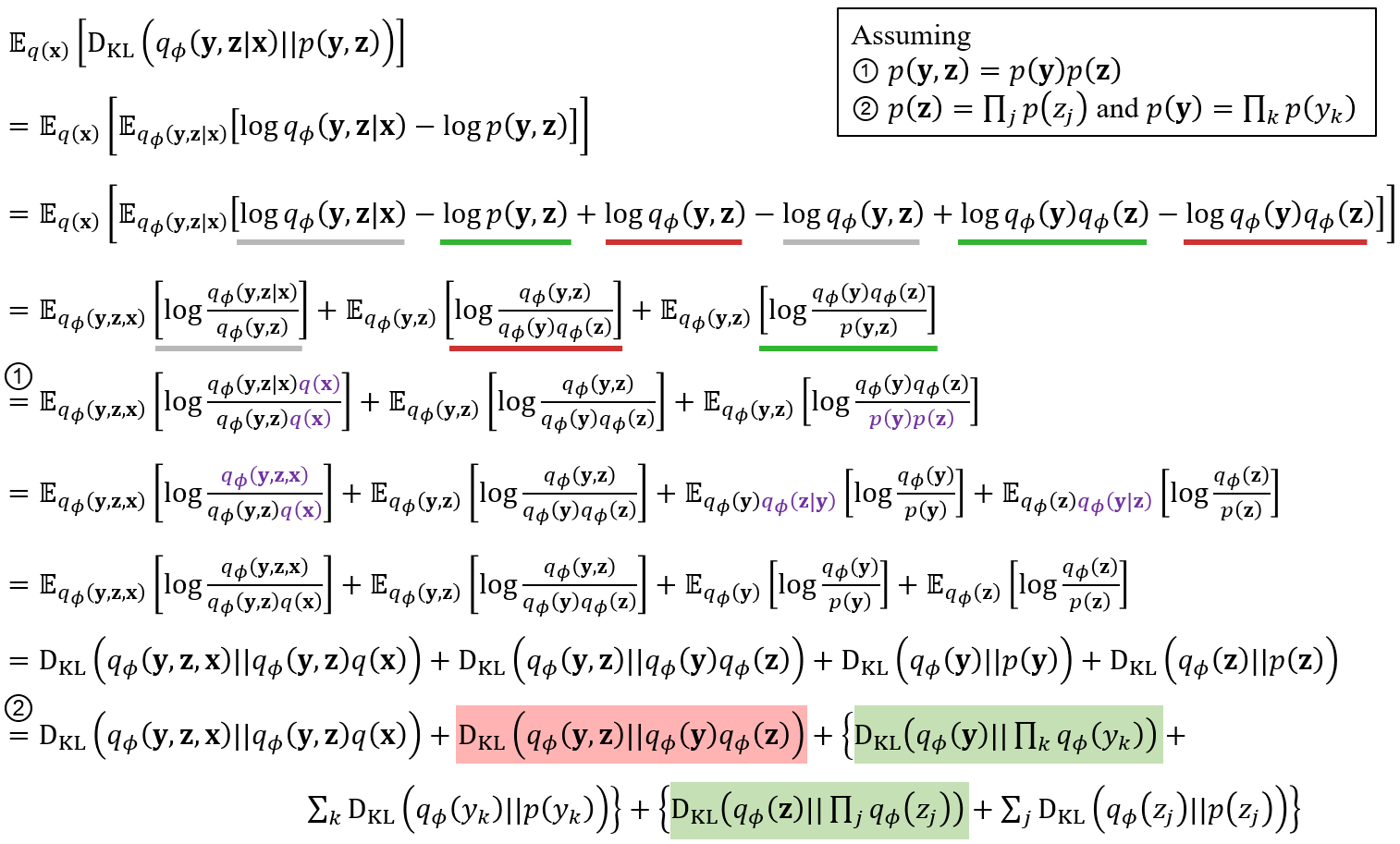}}
   \hypertarget{link_fig_derive_vecIdp}{\caption{Our decomposition of the KL term in the variational bound with two latent vectors. The terms for vector independence (highlighted in red) and variable independence (in green) are depicted.}}
 \label{fig_derive_vecIdp}
\end{center}
\end{figure}

\begin{figure}[]
\vskip -3cm
\begin{center}
\centerline{\includegraphics[width=13cm]{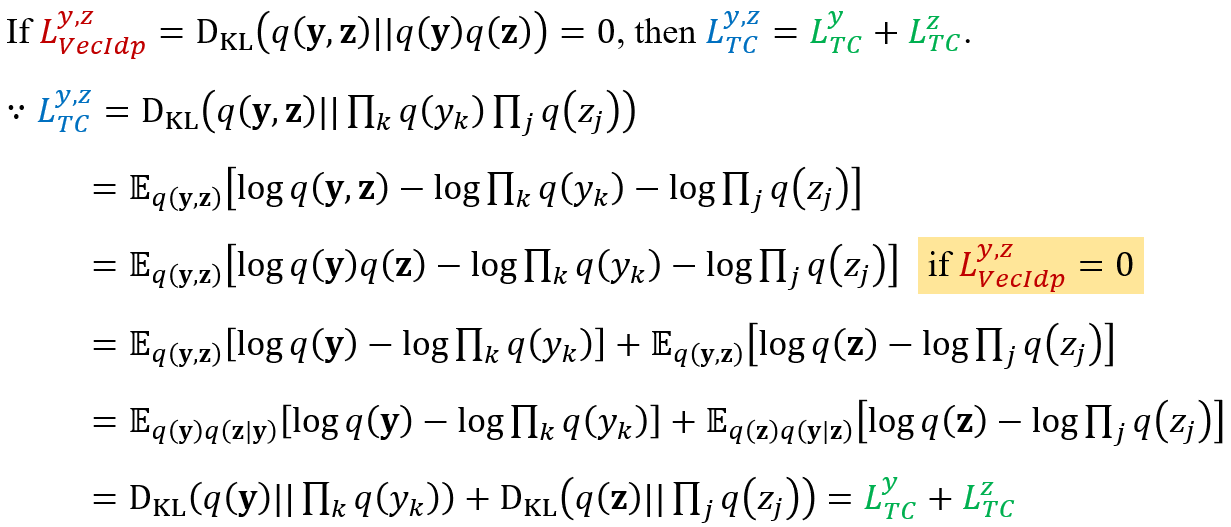}}
   \hypertarget{link_fig_derive_rel_vecIdp_tc}{\caption{Relationship between the vector independence objective and the separate TC: perfect vector independence ensures that the collective TC becomes the sum of the two separate TCs.}}
 \label{fig_derive_rel_vecIdp_tc}
\end{center}
\end{figure}


\hypertarget{supple_s4_expSetting}{\section{Data and Experimental Settings}}

We used the dSprites \cite{dsprites17}, MNIST \cite{lecun2010mnist}, and Fashion-MNIST \cite{xiao2017fashionmnist} datasets. For semi-supervised learning, the labeled data were selected to be distributed evenly across classes. The size of labeled set $L$ was either 2\% or 0.25\% of the entire training data, which we used for unlabeled set $U$. The classification loss weight, $\alpha=\rho\frac{N_L+N_U}{N_L}$, was set as $\alpha=51$ for the 2\% and 0.25\%-labeled dSprites setups. Here, for the 0.25\% setup, we failed to apply scaling constant $\rho=1$ (which caused an extremely large $\alpha$), and we instead used $\alpha=51$ as applied in the 2\% setup. We used $\alpha=5.1$ (i.e., $\rho=0.1$) for the 2\%-labeled MNIST. We used $\alpha=40.8$ (i.e., $\rho=0.8$) for the 2\%-labeled Fashion-MNIST.

Figure \hyperlink{link_fig_archi}{S3} shows the network architectures used in our experiments, which are identical to the convolutional architectures used in \cite{nips18_jointVae, icml19_cascadeVae}. The priors were set as $p(\textbf{z})=N(\textbf{0},\textbf{I})$ and $p(y)=Cat(\textbf{\textpi})$, where \textbf{\textpi} denotes evenly distributed class probabilities. We employed the Gumbel-Softmax distribution \cite{iclr17_gumbel, iclr17_concrete} for reparametrizing categorical $y$. We will release our experimental codes on GitHub.

\begin{figure}[t]
\begin{center}
\centerline{\includegraphics[width=16.3cm]{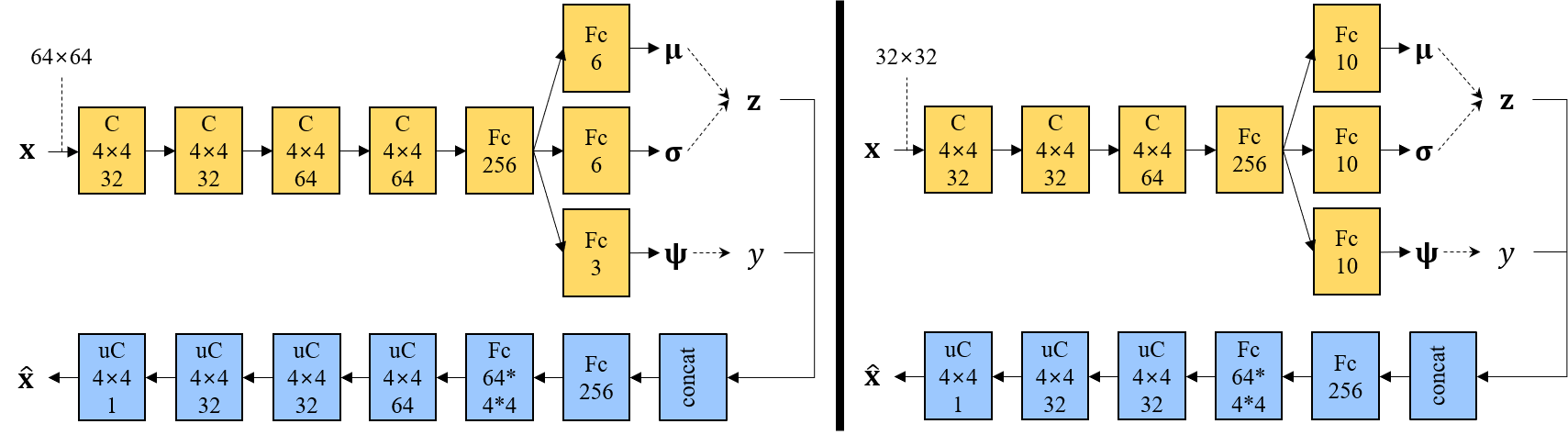}}
   \hypertarget{link_fig_archi}{\caption{Network architecture for 64×64 (left) and 32×32 (right) images. The yellow and blue boxes indicate the layers of encoder $\phi$ and decoder $\theta$, respectively. Each convolutional (C) or up-convolutional (uC) layer is identified by the size and number of filters. The stride size is 2 for the C and uC layers. Each fully-connected (Fc) layer is identified by the number of output neurons. The concatenation is denoted by “concat.” Each layer is followed by ReLU nonlinearity except that the last encoding layer and output layer are followed by linear and sigmoid activations, respectively. The dashed lines indicate the sampling operation for distributions $q_{\phi}(\textbf{z}|\textbf{x})= N(\textbf{\textmu},\textbf{\textsigma}^2\textbf{I})$ and $q_{\phi}(y |\textbf{x}) =Cat(\textbf{\textpsi})$.}}
 \label{fig_archi}
\end{center}
\vskip -0.5cm
\end{figure}

\subsection{dSprites: 3-class Shape Classification}

The dSprites dataset contains 2D-shape binary images with a size of 64$\times$64, which were synthetically generated with five independent factors: shape (3 classes; heart, oval, and square), position X (32 values), position Y (32), scale (6), and rotation (40). We divided the 737,280 images into training, validation, and test sets in a ratio of 10:1:1. We tested two cases with the 2\% and 0.25\% labels available in the training data. 

One 6-dimensional isotropic Gaussian vector was used for $q_{\phi}(\textbf{z}|\textbf{x})$, and one categorical variable representing 3 classes was used for $q_{\phi}(y|\textbf{x})$. The Gumbel-Softmax temperature parameter was set as 0.75. The Adam optimizer was used with an initial learning rate of 0.001 and a minibatch size of 2048. Every batch contained 1024 labeled samples. We trained networks for 100 epochs and reported the results measured at the epoch of the best validation loss. We tested 7 different random weeds for network weight initialization.

\subsection{MNIST: 10-class Digit Classification} 

The MNIST dataset contains 0--9 handwritten digit images with a size of 28$\times$28. We divided the original training set into 50,000 training and 10,000 validation images while maintaining the test set of 10,000 images. The images were normalized to have [0, 1] continuous values and resized to 32$\times$32 by following \cite{nips18_jointVae}. We tested the case with the 2\% labels available in the training data.

One 10-dimensional isotropic Gaussian vector was used for $q_{\phi}(\textbf{z}|\textbf{x})$, and one categorical variable representing 10 classes was used for $q_{\phi}(y|\textbf{x})$. The Gumbel-Softmax temperature parameter was set as 0.67. The Adam optimizer was used with an initial learning rate of 0.001 and a minibatch size of 512. Every batch contained 256 labeled samples. We trained networks for 200 epochs and reported the results measured at the epoch of the best validation loss. We tested 10 different random seeds for network weight initialization.

\subsection{Fashion-MNIST: 10-class Fashion Item Classification} 

The Fashion-MNIST dataset \cite{xiao2017fashionmnist} contains grayscale images with a size of 28$\times$28 and 10 fashion categories (e.g., t-shirt, trouser, sandal, and bag). The experimental settings were the same as those used in the MNIST dataset, except that the Gumbel-Softmax temperature parameter was set as 0.75.


\hypertarget{supple_s5_baselines}{\section{Baseline Methods}} \label{label_supple_s5_baselines}

We assumed that discrete random variable $y\in{\rm I\!R}$ and continuous random vector $\textbf{z}\in{\rm I\!R}^{J}$ were involved in the data generation process. Under this scenario, we compared our IV-VAE with the vanilla VAE \cite{iclr14_vae}, $\beta$-VAE \cite{iclr17_betaVae}, $\beta$-TCVAE \cite{nips18_betaTcVae}, and JointVAE \cite{nips18_jointVae}. For fair comparison, we applied the following settings that were used in our method to the baselines.

\vskip -0.4cm

\begin{itemize}[noitemsep]

\item We utilized the same dimensional latent units and network architectures. For the VAE, $\beta$-VAE, and $\beta$-TCVAE, we augmented the original continuous vector, $\textbf{z}$, in their objectives with discrete variable $y$. The jointVAE was originally designed to incorporate both of the continuous $\textbf{z}$ and discrete $y$.

\item We assumed the conditional independence of $q_{\phi}(y,\textbf{z}|\textbf{x})=q_{\phi}(y|\textbf{x})q_{\phi}(\textbf{z}|\textbf{x})$ by following \cite{nips18_jointVae}.

\item We modified their original unsupervised setups to use weak classification supervision for semi-supervised learning (SSL). Concretely, as applied in our method, their objectives for labeled set $L$ and unlabeled set $U$ became

\begin{center}
$\mathcal{J}^L (\phi,\theta) = \mathcal{J}_{recon}^L (\phi,\theta)+\mathcal{J}_{cls}(\phi)-\mathcal{J}^{Both} (\phi)$ \hspace{0.1cm} and \hspace{0.1cm} $\mathcal{J}^U (\phi,\theta) = \mathcal{J}_{recon}^U (\phi,\theta)- \mathcal{J}^{Both} (\phi)$.
\end{center}

The reconstruction and classification terms were also identical to those of our method, as shown below. For the reconstruction terms, true label $t$ for $L$ and inferred feature $y$ for $U$ were used to enable the decoder to better recognize one-hot class vectors.

\begin{center}
$\mathcal{J}_{recon}^L (\phi,\theta) = \mathbb{E}_{q^L(\textbf{x},t)} \Big[ \mathbb{E}_{q_{\phi}(\textbf{z}|\textbf{x})} \big[ \log p_{\theta}(\textbf{x}|t,\textbf{z}) \big] \Big]$ \hspace{0.1cm} and \hspace{0.1cm} $\mathcal{J}_{recon}^U (\phi,\theta) = \mathbb{E}_{q^U(\textbf{x})} \Big[ 
\mathbb{E}_{q_{\phi}(y,\textbf{z}|\textbf{x})} \big[ 
\log p_{\theta}(\textbf{x}|y,\textbf{z}) \big] \Big]$
\end{center}

\begin{center}
$\mathcal{J}_{cls} (\phi) = \alpha \, \mathbb{E}_{q^L(\textbf{x},t)} \Big[ 
\mathbb{I}(y=t) \log q_{\phi}(y|\textbf{x}) \Big]$
\end{center}

\end{itemize}

\vskip -0.15cm


The methods differ in the definition of $\mathcal{J}^{Both} (\phi)$ to learn disentangled features and regularize the encoder, as shown below. For notational brevity, we omit the dependence of objectives $\mathcal{J}$ and $\mathcal{L}$ on their parameters $\phi$. The terms related to independence of latent units are indicated by colors (i.e., $\textcolor{Green}{\mathcal{L}^{z}_{TC}}$, $\textcolor{blue}{\mathcal{L}^{y,z}_{TC}}$, and $\textcolor{RubineRed}{\mathcal{L}^{y,z}_{VecIdp} }$). Note that we did not penalize the latent-data MI terms by setting $\delta_y=\delta_z=\delta=0$ in order to help $y$ and $\textbf{z}$ capture the information about $\textbf{x}$.

\vskip -0.4cm


\begin{itemize}[noitemsep]

\item \textbf{VAE}: the case with $\beta=1$ in the below term of $\beta$-VAE

\item \textbf{$\beta$-VAE}: $\mathcal{J}^{Both}  = \beta \,\mathbb{E}_{q(\textbf{x})} \Big[ \textrm{D}_{\textrm{KL}} \big( q_{\phi}(y,\textbf{z} | \textbf{x}) || p(y,\textbf{z}) \big) \Big] = \beta \, \mathbb{E}_{q(\textbf{x})} \Big[ \textrm{D}_{\textrm{KL}}\big( q_{\phi}(y|\textbf{x}) || p(y)\big) + \textrm{D}_{\textrm{KL}}\big( q_{\phi}(\textbf{z}|\textbf{x}) || p(\textbf{z})\big)  \Big] $

\item \textbf{$\beta$-TCVAE-1}: $\mathcal{J}^{Both}  = \Big( \delta_y \, I_{q_{\phi}}(y;\textbf{x}) +  \gamma_y \, \mathcal{L}^y_{Reg} \Big)   + \Big( \delta_z \, I_{q_{\phi}}(\textbf{z};\textbf{x}) +\beta_z \, \textcolor{Green}{\mathcal{L}^{z}_{TC}} + \gamma_z \, \mathcal{L}^z_{Reg} \Big)  $
 
\item \textbf{$\beta$-TCVAE-2}: $\mathcal{J}^{Both} = \delta \, I_{q_{\phi}}(y,\textbf{z};\textbf{x}) +  \beta \, \textcolor{blue}{\mathcal{L}^{y,z}_{TC}} + \gamma_y \, \mathcal{L}^y_{Reg}  + \gamma_z \, \mathcal{L}^z_{Reg}  $

\item \textbf{JointVAE}: $\mathcal{J}^{Both}  = \beta \, \mathbb{E}_{q(\textbf{x})} \Big[ \left| \textrm{D}_{\textrm{KL}}\big(  q_{\phi}(y|\textbf{x}) || p(y)\big) - C_y \right|  \Big] + \beta \, \mathbb{E}_{q(\textbf{x})} \Big[ \left| \textrm{D}_{\textrm{KL}}\big( q_{\phi}(\textbf{z}|\textbf{x}) || p(\textbf{z})\big) - C_z \right| \Big] $

\item \textbf{IV-VAE-1 (ours)}: $\mathcal{J}^{Both}  =  
\delta \, I_{q_{\phi}}(y,\textbf{z};\textbf{x}) + \lambda \, \textcolor{RubineRed}{\mathcal{L}^{y,z}_{VecIdp} } +\gamma_y \, \mathcal{L}^y_{Reg}   + \beta_z \, \textcolor{Green}{\mathcal{L}^{z}_{TC}} + \gamma_z \, \mathcal{L}^z_{Reg} $

\item \textbf{IV-VAE-2 (ours)}: $\mathcal{J}^{Both}  =  
\delta \, I_{q_{\phi}}(y,\textbf{z};\textbf{x}) + \lambda \, \textcolor{RubineRed}{ \mathcal{L}^{y,z}_{VecIdp} } +\gamma_y \, \mathcal{L}^y_{Reg}   + \beta \, \textcolor{blue}{\mathcal{L}^{y,z}_{TC}} + \gamma_z \, \mathcal{L}^z_{Reg} $

\end{itemize}


\vskip -0.15cm

\noindent where $\mathcal{L}^{z}_{TC} = \textrm{D}_{\textrm{KL}} \big( q_{\phi}(\textbf{z}) || \prod\nolimits_{j} q_{\phi}(z_{j})\big) $ is the separate TC on $\textbf{z}$, $\mathcal{L}^{y,z}_{TC}= \textrm{D}_{\textrm{KL}} \big( q_{\phi}(y, \textbf{z}) || q_{\phi}(y) \prod\nolimits_{j}q_{\phi}(z_{j})\big)$ is the collective TC that simultaneously considers $y$ and the variables in $\textbf{z}$, $\mathcal{L}^{y,z}_{VecIdp} = \textrm{D}_{\textrm{KL}}\big( q_{\phi}(y,\textbf{z}) || q_{\phi}(y)q_{\phi}(\textbf{z})\big)$ is the vector independence term between $y$ and $\textbf{z}$, and $\mathcal{L}^y_{Reg} = \textrm{D}_{\textrm{KL}} \big( q_{\phi}(y)||p(y)\big)$ and $\mathcal{L}^z_{Reg} = \sum\nolimits_{j}\textrm{D}_{\textrm{KL}} \big( q_{\phi}(z_j)||p(z_j)\big)$ are the dimension-wise regularization terms. 

The $\mathcal{J}^{Both}$ term of \textbf{$\beta$-TCVAE-1 } is obtained by applying the KL decomposition of (4) in the main paper separately to $\mathbb{E}_{q(\textbf{x})} \big[ \textrm{D}_{\textrm{KL}}\big( q_{\phi}(y|\textbf{x}) || p(y)\big) \big]$ and $\mathbb{E}_{q(\textbf{x})} \big[ \textrm{D}_{\textrm{KL}}\big( q_{\phi}(\textbf{z}|\textbf{x}) || p(\textbf{z})\big)  \big]$ in the $\beta$-VAE objective and further assigning individual loss weights to the decomposed terms. Notice that $ \mathcal{L}^{y}_{TC}$ does not exist for the single-dimensional variable $y$ (i.e., $\mathcal{L}^{y}_{TC}= \textrm{D}_{\textrm{KL}} \big( q_{\phi}(\textbf{y}) || \prod\nolimits_{k=1}^{K} q_{\phi}(y_{k}) \big)$ with $K=1$ is zero). The $\mathcal{J}^{Both}$ term of \textbf{$\beta$-TCVAE-2 } is obtained by applying the KL decomposition of (4) in the main paper to $\mathbb{E}_{q(\textbf{x})} \big[ \textrm{D}_{\textrm{KL}} \big( q_{\phi}(y,\textbf{z} | \textbf{x}) || p(y,\textbf{z}) \big) \big]$ in the $\beta$-VAE objective, where the concatenation $[y;\textbf{z}]$ is considered as a single latent vector, and further assigning individual loss weights to the decomposed terms. For fair comparison, we assigned $\gamma_y$ and $\gamma_z$ separately to $\mathcal{L}^y_{Reg}$ and $\mathcal{L}^z_{Reg}$ instead of assigning a single $\gamma$ weight to $\mathcal{L}^y_{Reg} + \mathcal{L}^z_{Reg}$. The $\mathcal{J}^{Both}$ term of \textbf{JointVAE} is identical to the KL term in the original JointVAE objective, where the channel capacity weights $C_y$ and $C_z$ control the amount of information that $y$ and $\textbf{z}$ can capture. The $\mathcal{J}^{Both}$ term of \textbf{IV-VAE-1} is identical to (20) in the main paper. The $\mathcal{J}^{Both}$ term of \textbf{IV-VAE-2} is obtained by changing the TC term in the IV-VAE-1 objective from $\mathcal{L}^{z}_{TC}$ to $\mathcal{L}^{y,z}_{TC}$.

The $\beta$-TCVAE and IV-VAE in the main paper correspond to the $\beta$-TCVAE-1 and IV-VAE-1. In this document, we will present the additional results of $\beta$-TCVAE-2 and IV-VAE-2 along with $\beta$-TCVAE-1 and IV-VAE-1 in Section \hyperlink{supple_s6_dSprites}{S6}. We will analyze (i) the relationships between the vector independence and evaluation metrics and (2) the effect of penalizing collective TC $\mathcal{L}^{y,z}_{TC}$ or separate TC $\mathcal{L}^{z}_{TC}$ under encouraging vector independence term $\mathcal{L}^{y,z}_{VecIdp}$.


 \hypertarget{supple_s6_dSprites}{\section{In-depth Analysis of Relationships between Vector Independence and Evaluation Metrics} } \label{label_supple_s6_dSprites}

In the main paper, Figure 4 shows the results of $\beta$-TCVAE-1 and IV-VAE-1 on the dSprites dataset. Here, Figure \hyperlink{link_fig_metricRel_allTC}{S4} shows the results of $\beta$-TCVAE-2 and IV-VAE-2 (trained by penalizing the collective TC, $\mathcal{L}^{y,z}_{TC}$) that are merged to those of $\beta$-TCVAE-1 and IV-VAE-1 (trained by penalizing the separate TC, $\mathcal{L}^{z}_{TC}$). The x-axis shows vector independence KL $\mathcal{L}^{y,z}_{VecIdp} = \textrm{D}_{\textrm{KL}}\big( q_{\phi}(y,\textbf{z}) || q_{\phi}(y)q_{\phi}(\textbf{z})\big)$, where lower KL values indicate stronger independence. The y-axis shows (a) ELBO, (b) $\text{MIG}_{\text{all}}$, (c) $\text{MIG}_{\text{class}}$, (d) $\text{MIG}_{\text{style}}$, (e) classification error, (f) $I(y; t)$, and (g)$I(\textbf{z}; t)$ obtained under the SSL setups with 0.25\% and 2\% labeled data (denoted by SSL L0.25\% and SSL L2\%, respectively). Lower is better for (e) and (g), and higher is better for the other metrics. In Figure \hyperlink{link_fig_metricRel_allTC}{S4}-1 (left), each circle shows the median score of seven networks initiated from different random seeds. As shown in the legend, different colors indicate different $\beta_z$ and $\beta$ values for controlling the effect of TC in training, and bigger circles indicate bigger $\lambda$ values for causing stronger vector independence. In Figure \hyperlink{link_fig_metricRel_allTC}{S4}-2 (right), the different $\beta_z$ or $\beta$ settings are further merged to analyze the overall tendency, and each circle shows the median score of 21 networks (i.e., 7 random seeds $\times$ 3 $\beta_z$ or $\beta$ settings). 

We compare the effect of the collective TC on disentanglement learning with that of the separate TC. We also show the benefit of encouraging vector independence under penalizing not only the separate TC but also the collective TC. The observations are summarized below.

\vskip -0.8cm
\begin{itemize}[noitemsep]

\item In Figure \hyperlink{link_fig_metricRel_allTC}{S4}-1, given a larger value of $\beta=\beta_z$, penalizing the collective TC for $\beta$-TCVAE-2 and IV-VAE-2 yielded stronger vector independence than penalizing the separate TC for $\beta$-TCVAE-1 and IV-VAE-1, e.g., the results for $\beta=8$ (colored by cyan in Figure \hyperlink{link_fig_metricRel_allTC}{S4}-1) were placed to the left of those for $\beta_z=8$ (by blue), whereas the results for $\beta=1$ (by pink) and those for $\beta_z=1$ (by red) were similarly placed on the x-axis,. The cause of this phenomenon is that promoting independence between $y$ and all the variables in $\textbf{z}$ by decreasing the collective TC naturally encourages vector independence between $y$ and $\textbf{z}$ (e.g., independence between $y$, $z_1$, and $z_2$ ensures independence between $y$ and $\textbf{z}=[z_1, z_2]^T$). Nevertheless, increasing $\lambda$ for IV-VAE-2 under each $\beta$ setting often led to better vector independence than $\beta$-TCVAE-2 (i.e., $\lambda=0$). This phenomenon is caused by the imperfect minimization of the collective TC because of the existence of the other loss terms (e.g., the reconstruction term).

\vskip 0.15cm

\item In terms of the disentanglement scores, 
penalizing the separate TC $\mathcal{L}^{z}_{TC}$ on $\textbf{z}$ was competitive with or sometimes outperformed penalizing the collective TC $\mathcal{L}^{y,z}_{TC}$. For example, under the 0.25\% SSL setup in Figure \hyperlink{link_fig_metricRel_allTC}{S4}-2, the best scores of $\text{MIG}_{\text{all}}$, $\text{MIG}_{\text{class}}$, $\text{MIG}_{\text{style}}$, classification error, $I(y; t)$, and $I(\textbf{z}; t)$ obtained using the separate TC (text in orange) were better than those using the collective TC (text in purple). Rather than promoting multiple independences via the collective TC (e.g., $y \perp z_1$, $y \perp z_2$, and $z_1 \perp z_2$), promoting vector independence between $y$ and the entire $\textbf{z}$ (i.e., $y \perp \textbf{z}$) may guide the encoder better to separate discrete and continuous information. Under this guidance, penalizing the separate TC (e.g., $z_1 \perp z_2$) may help each variable capture each continuous factor.

\vskip 0.15cm

\item In Figure \hyperlink{link_fig_metricRel_allTC}{S4}-2, IV-VAE-2 networks outperformed or were competitive with $\beta$-TCVAE-2 networks for most scores, demonstrating the benefit of vector independence even under penalizing the collective TC. In particular, the $\lambda$ of 0.5 for the 0.25\% SSL setup worked well, as shown in the purple text in Figure \hyperlink{link_fig_metricRel_allTC}{S4}-2. Under the 2\% SSL setup, the best performing IV-VAE-2 networks yielded similar results with those of the best $\beta$-TCVAE-2 networks.

\end{itemize}


\hypertarget{supple_s7_fashMni}{\section{Additional Results on Fashion-MNIST} } \label{supple_s7_fashMni}

Figures \hyperlink{link_fig_fashMni_supple_corrEx}{S5}, \hyperlink{link_fig_fashMni_supple_genY_1clo}{S6},  \hyperlink{link_fig_fashMni_supple_genY_2item}{S7}, \hyperlink{link_fig_fashMni_supple_zTrav_1shape}{S8}, and \hyperlink{link_fig_fashMni_supple_zTrav_2bright}{S9} depict the qualitative results of the two networks trained on Fashion-MNIST without and with the vector independence objective (i.e., $\beta$-TCVAE-1 and IV-VAE-1). The networks were obtained from the same random seed and loss weights except the $\lambda$ weight: the $\lambda$ of 4 was used for training the IV-VAE-1, while the $\beta_z$ of 96, $\gamma_z$ of 1, and $\gamma_y$ of 2 were commonly used for both networks. 
  
To visually show the corrected class labels by encouraging vector independence, Figure \hyperlink{link_fig_fashMni_supple_corrEx}{S5} depicts reconstruction examples with estimated class information, i.e., inferred class probabilities or one-hot labels via the argmax operation. The results obtained from the class probabilities were often blurry, indicating that the inputs were confusing to be classified. Employing vector independence often corrects the classification results by enforcing $y$ to better store the class factor. Figures \hyperlink{link_fig_fashMni_supple_genY_1clo}{S6} and \hyperlink{link_fig_fashMni_supple_genY_2item}{S7}
 depict generated fashion images given input clothing styles (e.g., brightness and width). We extracted $\textbf{z}$ from the input and set $y$ as each of the one-hot item labels. Without promoting vector independence, the generation of non-clothing items (e.g., sneaker and bag) frequently failed. Encouraging vector independence better disentangled styles from class information, improving the synthesis controllability. The latent traversals in Figures \hyperlink{link_fig_fashMni_supple_zTrav_1shape}{S8}
 and \hyperlink{link_fig_fashMni_supple_zTrav_2bright}{S9} show that the two dimensions of $\textbf{z}$ captured the continuous factors corresponding to shapes and brightness.


\hypertarget{supple_s8_mnist}{\section{Additional Results on MNIST} } \label{supple_s8_mnist}

Figures \hyperlink{link_fig_digitMni_supple_corrEx}{S10}, \hyperlink{link_fig_digitMni_supple_genY}{S11}, and \hyperlink{link_fig_digitMni_supple_zTrav}{S12} depict the qualitative results of the two networks trained on MNIST without and with the vector independence objective (i.e., $\beta$-TCVAE-1 and IV-VAE-1). The networks were obtained from the same random seed and loss weights except the $\lambda$ weight: the $\lambda$ of 8 was used for training the IV-VAE-1, while the $\beta_z$ of 32, $\gamma_z$ of 1, and $\gamma_y$ of 2 were commonly used for both networks. 

Figure \hyperlink{link_fig_digitMni_supple_corrEx}{S10} shows corrected classification examples, which are visualized using the reconstruction outputs with estimated class information. Figure \hyperlink{link_fig_digitMni_supple_genY}{S11} depicts generation examples of 0--9 digit images given input styles. Encouraging vector independence better displays the digit styles (e.g., writing angle and line width) by helping $\textbf{z}$ effectively capture style information. In Figure \hyperlink{link_fig_digitMni_supple_zTrav}{S12}, the latent traversals on $\textbf{z}$ show the discovered continuous factors of variation.

\clearpage

\begin{figure}[]
\begin{center}
\centerline{\includegraphics[width=\columnwidth]{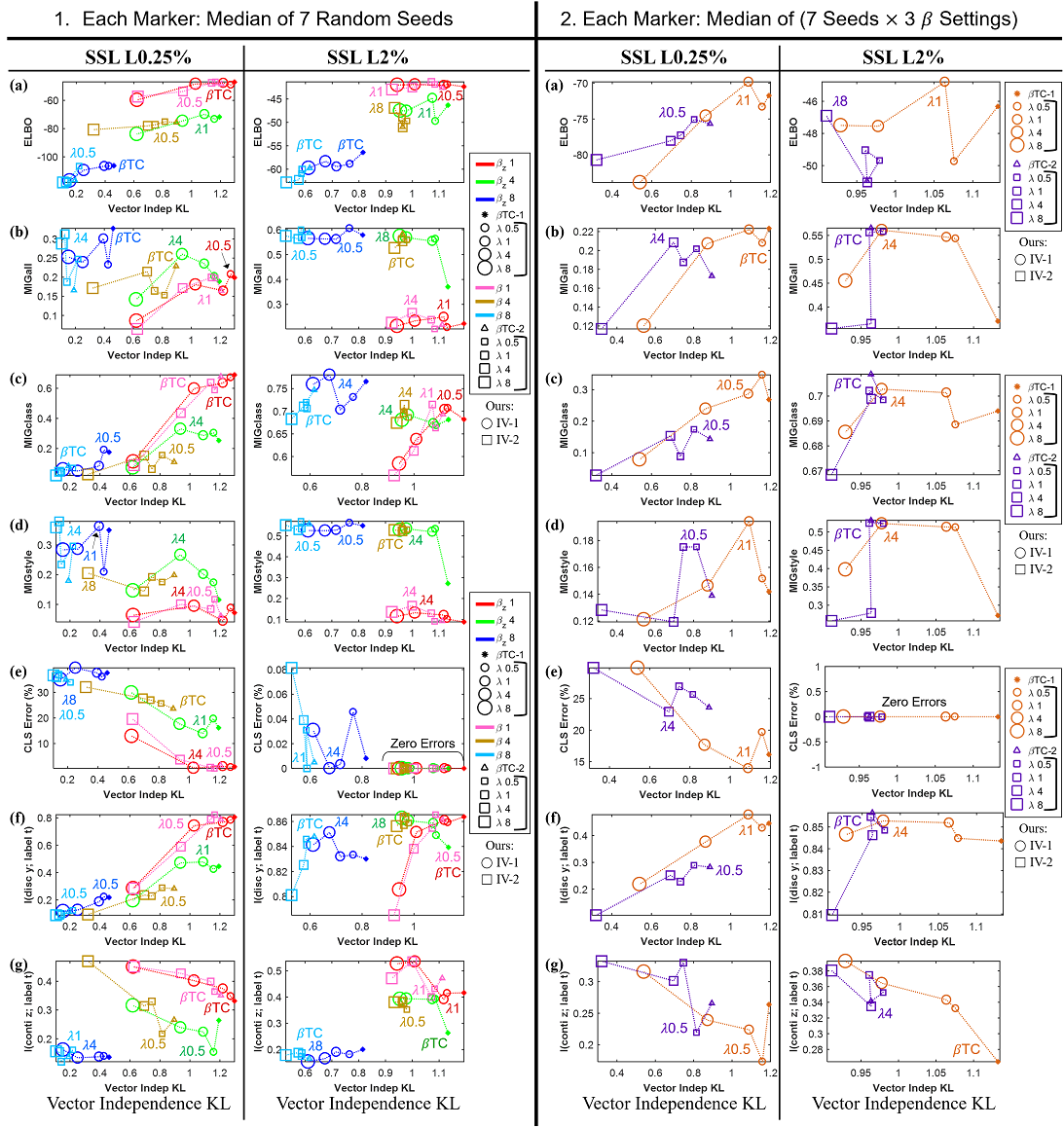}}
   \hypertarget{link_fig_metricRel_allTC}{\caption{Comparison of the collective TC and the separate TC. The $\beta$-TCVAE-2 and IV-VAE-2 were trained by penalizing the collective TC $\mathcal{L}^{y,z}_{TC}$ between $y$ and the variables in $\textbf{z}$, whereas the $\beta$-TCVAE-1 and IV-VAE-1 were trained by penalizing the separate TC $\mathcal{L}^{z}_{TC}$ on $\textbf{z}$. Given each setting, the best performing network is indicated with text. As shown in Figure \hyperlink{link_fig_metricRel_allTC}{S4}-1, given a larger value of $\beta_z=\beta$, penalizing the collective TC yields stronger vector independence than penalizing the separate TC. Nevertheless, because of the imperfect optimization of the collective TC, increasing $\lambda$ for IV-VAE-2 under each $\beta$ setting often yields better vector independence than $\beta$-TCVAE-2 (i.e., $\lambda=0$). As shown in Figure \hyperlink{link_fig_metricRel_allTC}{S4}-2, IV-VAE-2 outperforms or is competitive with $\beta$-TCVAE-2 for most scores, demonstrating the benefit of vector independence even under penalizing the collective TC.}}
 \label{fig_metricRel_allTC}
\end{center}
\end{figure}


\clearpage

\begin{figure}[]
\begin{center}
\centerline{\includegraphics[width=14cm]{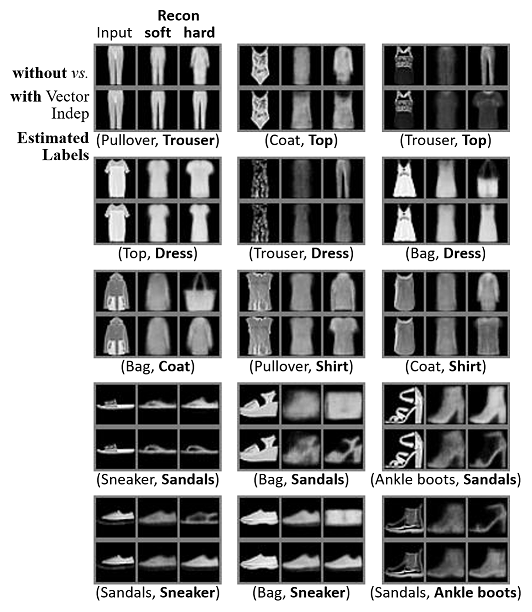}}
     \hypertarget{link_fig_fashMni_supple_corrEx}{\caption{Corrected classification examples on Fashion-MNIST. Given an input (left), inferred class probabilities (middle; denoted by “soft") or one-hot labels (right; denoted by “hard") from the class encoding $y$ are used for reconstruction. Below each example, inferred labels without and with the vector independence loss are indicated. Encouraging vector independence forces $y$ to better capture class information, improving the classification performance.}}
 \label{fig_fashMni_supple_corrEx}
\end{center}
\end{figure}

\clearpage

\begin{figure}[]
\begin{center}
\centerline{\includegraphics[width=13.1cm]{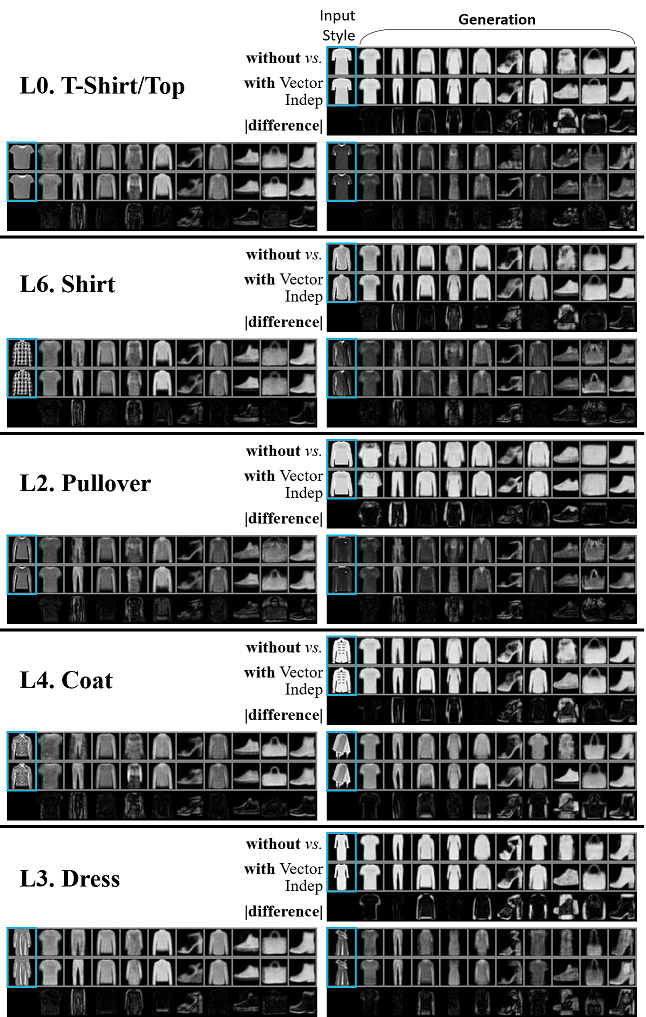}}
   \hypertarget{link_fig_fashMni_supple_genY_1clo}{\caption{Style-controlled generation on Fashion-MNIST. The style feature $\textbf{z}$ is extracted from an input (leftmost), and the class feature $y$ is set as desired item labels (right). Here, the style inputs correspond to \textit{top- and whole-body clothing} items, and their class number and name are indicated (e.g., L6. Shirt). The order of generation follows the order of original label numbers. Without promoting vector independence, the network often fails to generate Trouser (the 2nd column in the generation results), Sneaker (the 8th), and Bag (the 9th) classes. Encouraging vector independence forces $\textbf{z}$ to better capture style information by suitably separating styles from clothing classes.}}
 \label{fig_fashMni_supple_genY_1clo}
\end{center}
\end{figure}

\clearpage

\begin{figure}[]
\begin{center}
\centerline{\includegraphics[width=13.1cm]{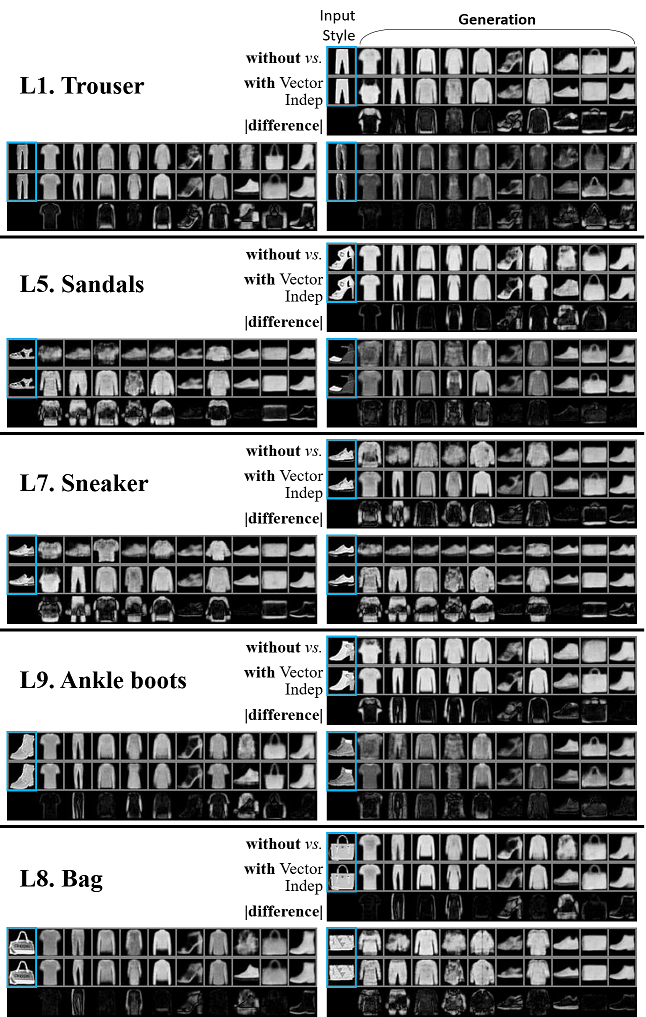}}
      \hypertarget{link_fig_fashMni_supple_genY_2item}{\caption{Style-controlled generation on Fashion-MNIST: continued from Figure \hyperlink{link_fig_fashMni_supple_genY_1clo}{S6}. The style feature $\textbf{z}$ is extracted from an input (leftmost), and the class feature $y$ is set as desired item labels (right). Here, the style inputs correspond to \textit{bottom clothing and non-clothing fashion} items, and their class number and name are indicated (e.g., L1. Trouser). The order of generation follows the order of original label numbers. Without promoting vector independence, the generation with shoe styles frequently fails, as shown in the results of L5, L7, and L9. Encouraging vector independence forces $\textbf{z}$ to better capture style information by suitably separating styles from clothing classes.}}
 \label{fig_fashMni_supple_genY_2item}
\end{center}
\end{figure}

\clearpage

\begin{figure}[]
\begin{center}
\centerline{\includegraphics[width=16cm]{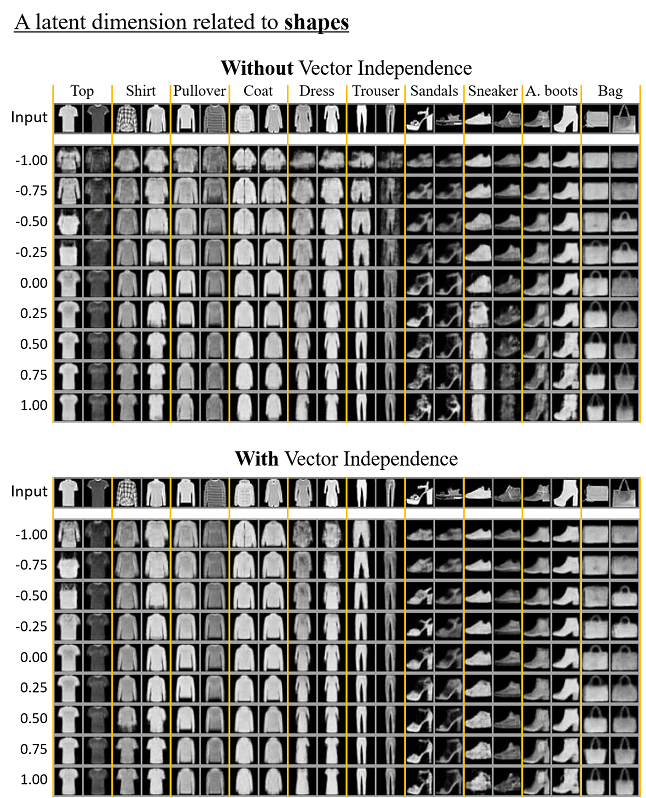}}
   \hypertarget{link_fig_fashMni_supple_zTrav_1shape}{\caption{Latent traversal examples on Fashion-MNIST. The class feature $y$ and style feature $\textbf{z}$ are extracted from the given inputs (top-most), and the latent traversal results on a single dimension $z_j$ are depicted (bottom; traversal range is from -1 to 1). The models capture the continuous factor related to clothing \textit{shapes} (e.g., width, sleeve length, heel height, and bag size).}}
 \label{fig_fashMni_supple_zTrav_1shape}
\end{center}
\end{figure}

\clearpage

\begin{figure}[]
\begin{center}
\centerline{\includegraphics[width=16cm]{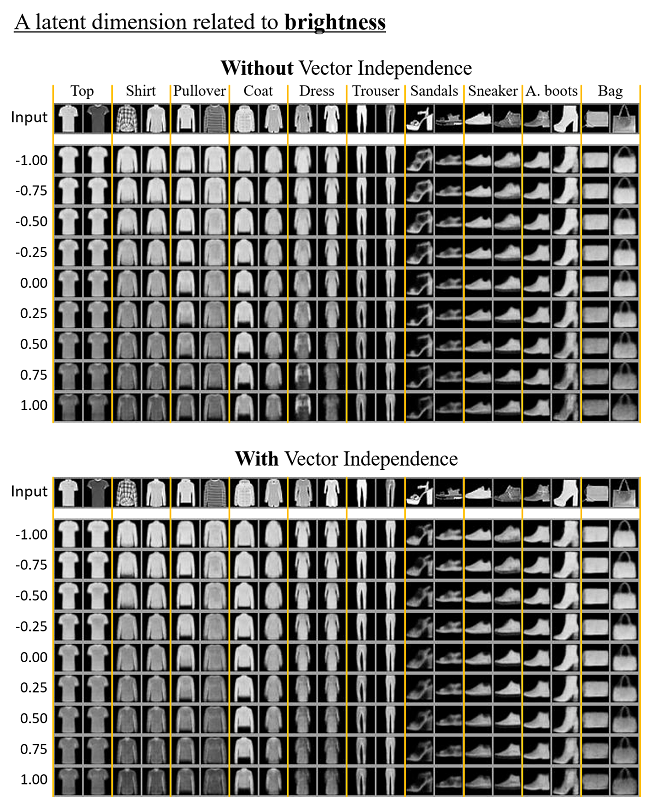}}
         \hypertarget{link_fig_fashMni_supple_zTrav_2bright}{\caption{Latent traversal examples on Fashion-MNIST: continued from Figure \hyperlink{link_fig_fashMni_supple_zTrav_1shape}{S8}
. The class feature $y$ and style feature $\textbf{z}$ are extracted from the given inputs (top-most), and the latent traversal results on a single dimension $z_j$ are depicted (bottom; traversal range is from -1 to 1). The models capture the continuous factor related to \textit{brightness}.}}
 \label{fig_fashMni_supple_zTrav_2bright}
\end{center}
\end{figure}


\clearpage

\begin{figure}[]
\begin{center}
\centerline{\includegraphics[width=15.2cm]{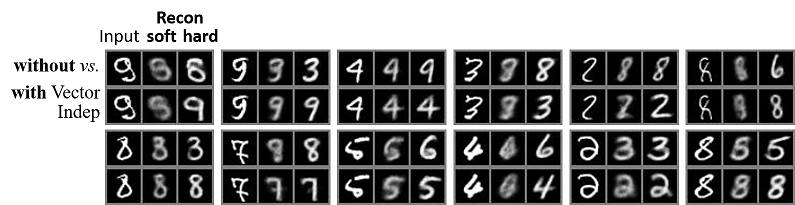}}
  \hypertarget{link_fig_digitMni_supple_corrEx}{ \caption{Corrected classification examples on MNIST. Given an input (left), inferred class probabilities (middle; denoted by “soft") or one-hot labels (right; denoted by “hard") from the class encoding $y$ are used for reconstruction. Encouraging vector independence forces $y$ to better capture class information, improving the classification performance.}}
 \label{fig_digitMni_supple_corrEx}
\end{center}
\end{figure}

\vskip -3cm

\begin{figure}[]
\begin{center}
\centerline{\includegraphics[width=14.2cm]{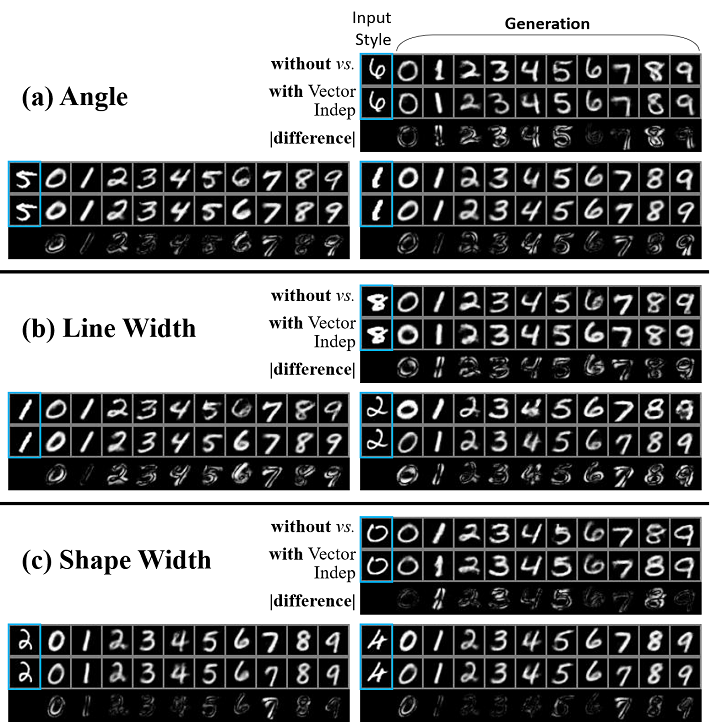}}
   \hypertarget{link_fig_digitMni_supple_genY}{ \caption{Style-controlled generation on MNIST. The style feature $\textbf{z}$ is extracted from an input (leftmost), and the class feature $y$ is set as desired digit labels (right). The order of generation follows the order of original label numbers. Encouraging vector independence forces $\textbf{z}$ to better capture digit styles such as (a) writing angle, (b) line width, and (c) shape width.}}
 \label{fig_digitMni_supple_genY}
\end{center}
\end{figure}

\clearpage

\begin{figure}[]
\begin{center}
\centerline{\includegraphics[width=15cm]{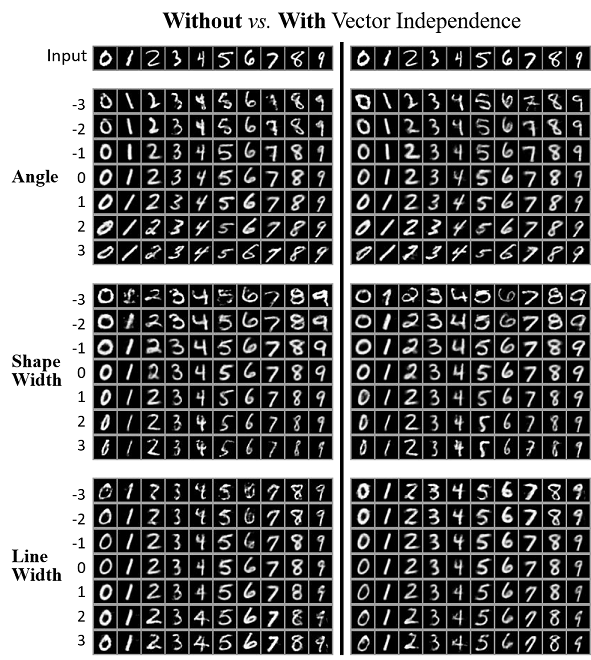}}
   \hypertarget{link_fig_digitMni_supple_zTrav}{\caption{Latent traversal examples on MNIST. The class feature $y$ and style feature $\textbf{z}$ are extracted from the given inputs (top-most), and the latent traversal results on the three dimensions of $\textbf{z}$ are depicted (bottom; traversal range is from -3 to 3). The continuous factors related to digit styles are stored in $\textbf{z}$.} }
 \label{fig_digitMni_supple_zTrav}
\end{center}
\end{figure}

\clearpage


\end{document}